\pdfoutput=1
\documentclass[11pt]{article}

\usepackage[]{acl}

\usepackage{times}
\usepackage{latexsym}

\usepackage[T1]{fontenc}
\usepackage[utf8]{inputenc}

\usepackage{microtype}

\usepackage{amsmath}
\usepackage{amssymb}
\usepackage{hyperref}
\usepackage{url}
\usepackage{graphicx}
\usepackage{booktabs}
\usepackage{multirow}
\usepackage{array}
\usepackage{makecell}
\newcolumntype{P}[1]{>{\centering\arraybackslash}p{#1}}
\usepackage{tablefootnote}
\usepackage{threeparttable}

\usepackage{soul}
\definecolor{delectricorange}{RGB}{212,103,27}
\colorlet{lightdelectricorange}{delectricorange!30}
\newcommand{\hldbOrange}[1]{%
    {%
    \sethlcolor{lightdelectricorange}%
    \hl{#1}%
    }%
}
\definecolor{delectricgreen}{RGB}{97,150,64}
\colorlet{lightdelectricgreen}{delectricgreen!30}
\newcommand{\hldbGreen}[1]{%
    {%
    \sethlcolor{lightdelectricgreen}%
    \hl{#1}%
    }%
}

\newcommand{\Blue}[1]{\textcolor[rgb]{0.00,0.00,0.00}{#1}}

\newcommand{\figGreen}[1]{\textcolor[RGB]{97,150,64}{#1}}
\newcommand{\figOrange}[1]{\textcolor[RGB]{212,103,27}{#1}}

\newcommand{\ourtask}[0]{\textit{Counterfactual Logical Modification}}
\newcommand{\ourdataset}[0]{\textsc{CLoMo}}
\newcommand{\argprime}[0]{\textsf{Argument$^{\prime}$}}
\newcommand{\argm}[0]{\textsf{Argument}}
\newcommand{\pone}[0]{\textsf{Premise1}}
\newcommand{\ptwo}[0]{\textsf{Premise2}}

\title{
\ourdataset: Counterfactual Logical Modification \\ with Large Language Models
}

\author{
    Yinya Huang\textsuperscript{1,8}\thanks{\quad These authors contributed equally to this work. Work is done during R. Hong's internship at Tencent AI Lab.} ~~~
    Ruixin Hong\textsuperscript{2}\footnotemark[1] ~~~
    Hongming Zhang\textsuperscript{3} ~~~ 
    Wei Shao\textsuperscript{1} ~~~ 
    Zhicheng Yang\textsuperscript{4} ~~~ \\
    {\bf Dong Yu\textsuperscript{3} ~~~}
    {\bf Changshui Zhang\textsuperscript{2} ~~~}
    {\bf Xiaodan Liang\textsuperscript{5,6,7}\thanks{\quad Corresponding author.} ~~~}
    {\bf Linqi Song\textsuperscript{1,8}}$^{\dag}$ \\
    $^1$City University of Hong Kong ~~
    $^2$Tsinghua University \\
    $^3$Tencent AI Lab, Seattle ~~
    $^4$The Hong Kong University of Science and Technology (Guangzhou) \\
    $^5$Shenzhen Campus of Sun Yat-sen University ~~
    $^6$MBZUAI ~~
    $^7$DarkMatter AI Research \\
    $^8$City University of Hong Kong Shenzhen Research Institute \\
    \tt yinya.huang@hotmail.com,
    \tt hrx20@mails.tsinghua.edu.cn
}
 
\begin{document}
\maketitle
\begin{abstract}
In this study, we delve into the realm of counterfactual reasoning capabilities of large language models (LLMs).
Our primary objective is to cultivate the counterfactual thought processes within LLMs and rigorously assess these processes for their validity. 
Specifically, we introduce a novel task, \underline{\textbf{C}}ounterfactual \underline{\textbf{Lo}}gical \underline{\textbf{Mo}}dification (\ourdataset), and a high-quality human-annotated benchmark. In this task, LLMs must adeptly alter a given argumentative text to uphold a predetermined logical relationship. 
To effectively evaluate a generation model's counterfactual capabilities, we propose an innovative evaluation metric, the 
% \textit{Logic-Aware Counterfactual Score} 
decomposed \textit{Self-Evaluation Score (SES)}
to directly evaluate the natural language output of LLMs instead of modeling the task as a multiple-choice problem.
Analysis shows that the proposed automatic metric aligns well with human preference.
Our experimental results show that while LLMs demonstrate a notable capacity for logical counterfactual thinking, there remains a discernible gap between their current abilities and human performance. 
Code and data are available at \url{https://github.com/Eleanor-H/CLOMO}.
\end{abstract}

\section{Introduction}
\label{sec:intro}

% fig 1
\begin{figure*}[!t]
\centering
    \vspace{-15pt}
    \includegraphics[width=\textwidth]{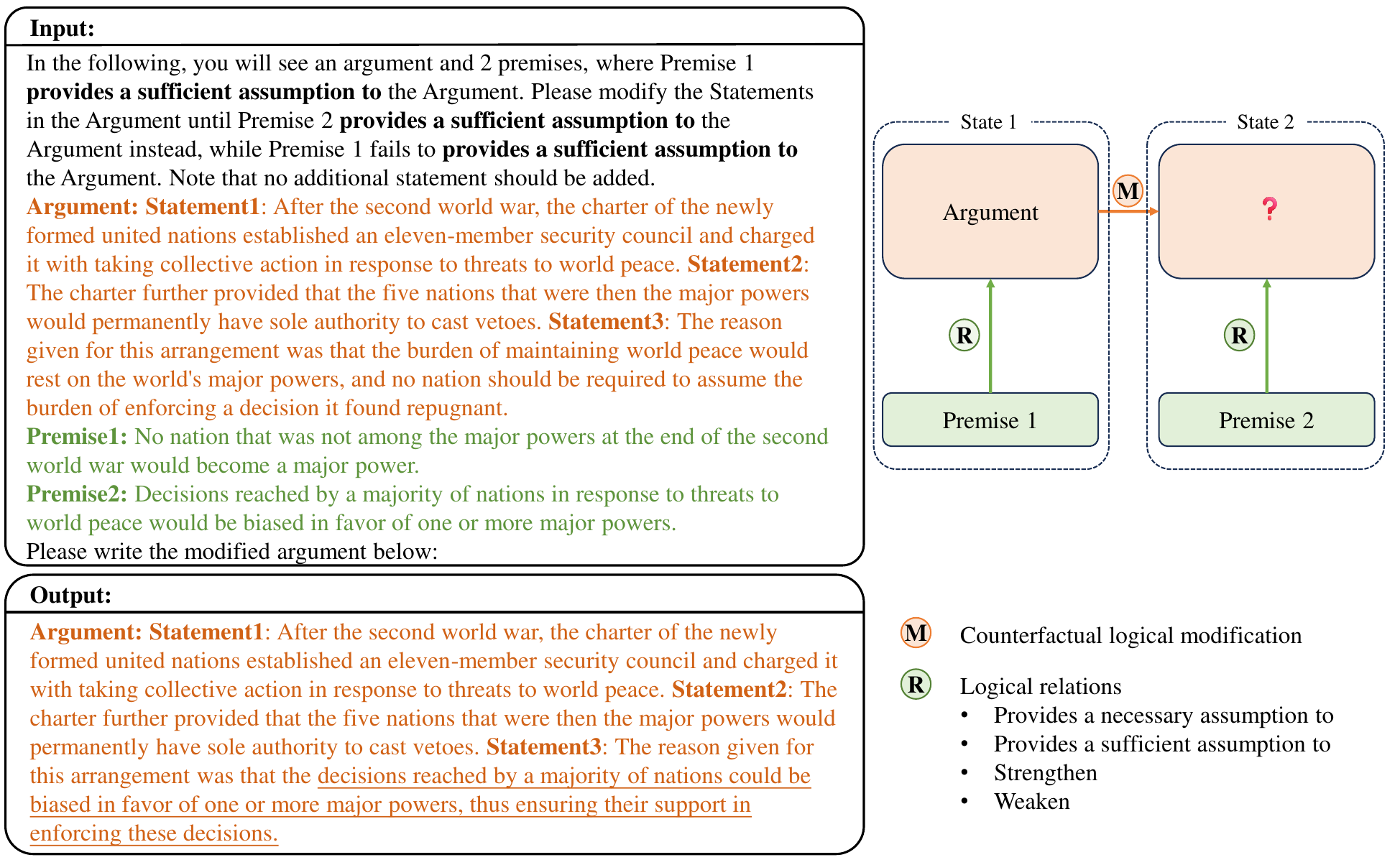}
    \vspace{-23pt}
    \caption{
        Demonstration of \ourdataset.
        An LLM is given an argument and two premises. The LLM needs to modify the statements in Argument such that the logical relation \textsf{R} switch to stand in \textsf{state 2} instead of \textsf{state 1}.}
    \vspace{-12pt}
    \label{fig:task}
\end{figure*}

% para 1
Despite large language models \cite{DBLP:journals/corr/abs-2308-03762,gpt35} perform strikingly in plenty of reasoning benchmarks \cite{DBLP:journals/corr/abs-2110-14168,DBLP:conf/iclr/HendrycksBBZMSS21},
late studies observe an internal inconsistency in their reasoning processes \cite{DBLP:conf/iclr/Saparov023,DBLP:journals/corr/abs-2308-03762}.
The inconsistency is attributed to misunderstanding and misapplication of logical relations. 
However, logical relations in complex language reasoning are not yet properly quantified and evaluated. 

% para 2
Current studies on evaluating model reasoning are limited in both form and content. 
On the one hand, benchmarking complex reasoning is generally applying discrimination tasks such as multiple-choice questions \cite{DBLP:journals/corr/abs-2305-08322,DBLP:conf/iclr/HendrycksBBZMSS21,DBLP:journals/corr/abs-2305-12524,DBLP:conf/acl/SuzgunSSGTCCLCZ23}, where accuracy and pass rate serve as the main evaluation metric.
However, such evaluations over-simplify the goal of uncovering essential and subtle pitfalls in complex reasoning.
For example, the reasoning processes could contain misconceptions in logical relations but give correct answers due to the data distribution~\cite{DBLP:conf/emnlp/ElazarZGR21,DBLP:conf/iclr/Saparov023}.
% the misconceptions in logical relations or pitfalls in the reasoning processes can be hardly observed.
% Moreover, it is observed that the generated chains of thought and the final answers can be uncorrelated \cite{DBLP:conf/iclr/Saparov023}, and a correct final answer does not guarantee a proper reasoning process. 
 % reduce the credibility of the accuracies. 
% Moreover, the generated chain of thoughts and the final answers can be inconsistent \cite{DBLP:conf/iclr/Saparov023}. 
Therefore, evaluating the generated content would provide a more realistic measurement of model reasoning. 
On the other hand, unlike widely studied reasoning tasks such as math reasoning \cite{DBLP:journals/corr/abs-2110-14168,DBLP:conf/nips/HendrycksBKABTS21} and standard exams \cite{DBLP:journals/corr/abs-2303-08774,DBLP:journals/corr/abs-2305-08322}, counterfactual reasoning \cite{sep-counterfactuals} as a fundamental evaluation of logical relations is less explored in the context of large language models. 
Previous literature studies counterfactual reasoning either in a multiple-choice manner \cite{DBLP:conf/emnlp/TandonDSCB19,DBLP:conf/acl/QinGUHCF20} or applying labored human study to evaluate counterfactual generation \cite{DBLP:conf/emnlp/QinBHBCC19}, leaving an effective evaluation of counterfactual generation unexplored. 

In our study, we delve into the realm of evaluating large language models' (LLMs) ability to generate counterfactually coherent thoughts. 
\Blue{Figure~\ref{fig:task} demonstrates the paradigm.}
Specifically, we proposed an innovative evaluation system that quantitatively measures the evolution of information in statement pairs, ensuring that they adhere to a specified logical relationship. Our approach includes designing a specialized task where models are presented with mismatched argument-premise pairs bound by a specific logical relation. The objective for these models is to adeptly modify the argument text until the specified logical relation is satisfactorily established. In conjunction with this task, we have created the first dataset of its kind, comprising dual argument-premise pairs, each annotated with a defined logical relation. This dataset is vital for facilitating logically restricted counterfactual modifications, and we have enriched it with human-written modifications to serve as a benchmark for evaluation.

% para 4: We treat the LLMs as a player (?) and complete the counterfactual modification game as a human does. experiments & findings. 
% We perform experiments with large language models including GPT-4 \cite{DBLP:journals/corr/abs-2303-08774} and GPT-3.5-Turbo \cite{gpt35} and small language models including the LLaMA and LLaMA 2 families. 
% We find that \ourdataset\ is a very challenging task, and the counterfactual logical reasoning of existing models needs to be improved.
Our experimental investigations encompass a range of large language models, including the latest GPT-4o\footnote{\url{https://openai.com/index/hello-gpt-4o/}}, GPT-4 \cite{DBLP:journals/corr/abs-2303-08774}, and GPT-3.5-Turbo \cite{gpt35}, as well as smaller models from the LLaMA~\cite{DBLP:journals/corr/abs-2302-13971} and LLaMA 2~\cite{DBLP:journals/corr/abs-2307-09288} families. Through these experiments, we have discerned that the task of \ourdataset\ poses a significant challenge. It becomes evident that these models' current counterfactual logical reasoning capabilities fall short of the desired proficiency. This observation underscores the need for further advancements in enhancing the counterfactual reasoning abilities of existing language models, paving the way for more sophisticated and logically coherent AI systems.

% para 5
% We summarize the contributions of this paper as follows:
The contributions of this paper are three-fold:
\vspace{-5pt}
\begin{itemize}\setlength\itemsep{-5pt}
\item We propose the task of \ourtask\ and contribute a corresponding \ourdataset
to evaluate the counterfactual reasoning capability of LLMs in the scenario of complicated textual logical reasoning. 
\item We propose the 
% \textit{Logic-Aware Counterfactual Score} 
decomposed \textit{Self-Evaluation Score (SES)}
for the logically consistent generation of large language models. 
\item We conduct experiments on LLMs (GPT-3.5, GPT-4) and small language models (the LLaMA and LLaMA 2 families)
and find that \ourdataset\ is a very challenging task and the counterfactual logical reasoning ability of the existing model needs to be improved.
\end{itemize}

\vspace{-7pt}
% table 1
\begin{table*}[!t]
    % \centering
    \scriptsize
    \begin{tabular}{
        P{.077\textwidth}
        >{\raggedright\arraybackslash}P{.42\textwidth}
        >{\raggedright\arraybackslash}P{.42\textwidth}
    }
        \toprule
        \textbf{\textsf{R}} & \textbf{\textsf{State 1}} & \textbf{\textsf{State 2}} \\
        \midrule
        \textit{Necessary Assumption (38.5\%)}
        & 
        \textbf{Argument:} 
        Statement1: Journalist: the advice of social scientists is frequently overlooked by politicians making social policy. Statement 2: Because it is not unreasonable to discount scientific assertions \hldbOrange{backed by weak evidence,} politicians should not generally be criticized for ignoring social science\hldbOrange{,} for social scientists, unlike physical scientists, seldom agree on the claims made even within their own specialty. \newline
        \hldbGreen{\textbf{Premise1:}} The failure of scientists to agree that a claim within their specialty is true can indicate that the evidence for the claim is not strong.
        & 
       \textbf{Argument$^{\prime}$:} 
       Statement 1: Journalist: the advice of social scientists is frequently overlooked by politicians making social policy. Statement 2: Because it is not unreasonable to discount scientific assertions\hldbOrange{, }politicians should not generally be criticized for ignoring social science \hldbOrange{unless social scientists agree on the same claim,} for social scientists, unlike physical scientists, seldom agree on the claims made even within their own specialty. \newline
       \hldbGreen{\textbf{Premise2:}} Politicians should follow the advice of experts on issues about which those experts agree among themselves.
        \\
        \midrule
        \textit{Sufficient Assumption (6.6\%)} 
        & 
        \textbf{Argument:} 
        Statement 1: Caffeine\hldbOrange{ }can kill or inhibit the growth of the larvae of several species of insects. Statement 2: One recent experiment showed that tobacco hornworm larvae die when they ingest a preparation that consists, in part, of finely powdered tea leaves, which contain caffeine. Statement 3: This result is evidence for the hypothesis that the presence of non-negligible quantities of caffeine in various parts of many diverse species of \hldbOrange{plants} is not accidental but evolved as a defense for \hldbOrange{those plants}. \newline
        \hldbGreen{\textbf{Premise1:}} Caffeine-producing plants or their ancestors have sometimes been fed upon by creatures sensitive to caffeine.
        &
        \textbf{Argument$^{\prime}$:} 
        Statement 1: Caffeine \hldbOrange{produced for plant species' own} \hldbOrange{defense} can kill or inhibit the growth of the larvae of several species of insects. Statement 2: One recent experiment showed that tobacco hornworm larvae die when they ingest a preparation that consists, in part, of finely powdered tea leaves, which contain caffeine. Statement 3: This result is evidence for the hypothesis that the presence of non-negligible quantities of caffeine in various parts of \hldbOrange{tobacco plant} is not accidental but evolved as a defense for \hldbOrange{it}. \newline
        \hldbGreen{\textbf{Premise2:}} The tobacco plant is among the plant species that produce caffeine for their own defense.
        \\
        \midrule
        \textit{Strengthen (18.7\%)} 
        &
        \textbf{Argument:} 
        Statement1: In contemplating major purchases, businesses often consider only whether there is enough money left from monthly revenues after paying monthly expenses to cover the cost of the purchase. But \hldbOrange{many expenses do not occur monthly} ; taking into account \hldbOrange{only monthly expenses} can cause a business to overexpand. Statement2: So the use of a cash-flow statement is critical for all businesses. \newline
        \hldbGreen{\textbf{Premise1:}} A cash-flow statement is the only way to track both monthly expenses and expenses that are not monthly.
        &
        \textbf{Argument$^{\prime}$:} 
        Statement 1: In contemplating major purchases, businesses often consider only whether there is enough money left from monthly revenues after paying monthly expenses to cover the cost of the purchase. But \hldbOrange{there are many expenses every month} ; taking into account \hldbOrange{these expenses incorrectly} can cause a business to overexpand. Statement 2: So the use of a cash-flow statement is critical for all businesses. \newline
        \hldbGreen{\textbf{Premise2:}} Only a cash-flow statement can accurately document all monthly expenses.
        \\
        \midrule
        \textit{Weaken (36.2\%)} 
        & 
        \textbf{Argument:} 
        Statement1: The country of baurisia has, until now, been self-sufficient in both grain and meat. However, with growing prosperity in baurisia has come a steadily increasing per capita consumption of meat\hldbOrange{, and it takes several pounds of grain to produce one pound of meat.} Statement2: Therefore, since per capita income in baurisia is almost certain to rise further but increases in domestic \hldbOrange{grain} production are highly unlikely, baurisia is soon likely to become an importer of \hldbOrange{grain}. \newline
        \hldbGreen{\textbf{Premise1:}} It is more economical for baurisians to import meat than grain.
        &
        \textbf{Argument$^{\prime}$:} 
        Statement 1: The country of baurisia has, until now, been self-sufficient in both grain and meat. However, with growing prosperity in baurisia has come a steadily increasing per capita consumption of meat\hldbOrange{.} Statement 2: Therefore, since per capita income in baurisia is almost certain to rise further but increases in domestic \hldbOrange{meat} production are highly unlikely, baurisia is soon likely to become an importer of \hldbOrange{meat}. \newline
        \hldbGreen{\textbf{Premise2:}} The per capita consumption of meat in baurisia is roughly the same across all income levels.
        \\
        \bottomrule
    \end{tabular}
    \vspace{-2pt}
    \caption{Example questions from the \ourdataset\ benchmark, with the proportion of each logical relation.
    \hldbOrange{Counterfactual logical modifications} regarding the \hldbGreen{change of state by a premise} are highlighted.
    }
    \vspace{-12pt}
    \label{tab:example}
\end{table*}

\vspace{-1pt}
\section{Related Works}
\vspace{-4pt}
\label{sec:related}
%
% \vspace{-5pt}
\paragraph{From Complex Reasoning to Counterfactual Reasoning}
% \vspace{-5pt}
% \paragraph{Complex Reasoning and Counterfactual Reasoning}
Complex reasoning has been highly concerned as a significant yet challenging task for inspecting advanced artificial intelligence. 
For example, for solving commonsense reasoning problems \cite{DBLP:conf/naacl/TalmorHLB19,DBLP:conf/nips/TalmorYBBGCB21,huang-etal-2019-cosmos,Bhagavatula2020Abductive,sap-etal-2019-social}, the models \cite{DBLP:journals/corr/abs-2104-06378,yasunaga2022deep,liu2021kg,huang2021rem} are required to reasonable applying commonsense knowledge to conduct the reasoning process for the final answer. 
Furthermore, multi-step reasoning needs the models to perform multiple reasoning steps while maintaining consistency and faithfulness. 
To achieve this, synthetic compositional reasoning tasks \cite{DBLP:journals/corr/abs-2009-07185,DBLP:conf/acl/TafjordDC21,han2022folio,DBLP:conf/iclr/Saparov023,huang2024mustard} incorporate first-order logic to inspect and improve models logical consistency \cite{DBLP:journals/corr/abs-2305-12295,DBLP:journals/corr/abs-2310-15164,DBLP:conf/acl/SanyalS022,DBLP:journals/corr/abs-2111-12038}. 
Moreover, real-scenario compositional reasoning \cite{yu2020reclor, DBLP:conf/ijcai/LiuCLHWZ20,DBLP:conf/emnlp/DalviJTXSPC21, DBLP:conf/emnlp/HuangZHLZ022} joins commonsense, consider the uncertainty of events in multi-step logical reasoning, which challenge current models \cite{bao2023enhancing,DBLP:conf/acl/Xu0CW23,DBLP:journals/corr/abs-2305-13718,DBLP:conf/acl/JiaoGSN22,DBLP:journals/pami/HuangLXFLL23,DBLP:conf/naacl/HuangFCWL21} to solve real-world reasoning problems with faithfulness. 
Additionally, the more faithful models should be able to consider counterfactuals.
For example, answering questions given counterfactual conditions \cite{DBLP:journals/corr/abs-2305-14010,DBLP:conf/emnlp/TandonDSCB19,DBLP:conf/acl/QinGUHCF20}, or narrating a counterfactual scenario \cite{DBLP:conf/emnlp/QinBHBCC19}.
However, previous studies on counterfactual reasoning barely pay attention to the logical consistency or faithfulness of models. 
Therefore, in this work, we propose \ourtask\ that challenges the model to satisfy a given logical relation restriction while generating counterfactuals.

\vspace{-5pt}
\paragraph{Evaluation of LLM Reasoning}
% \vspace{-5pt}
% Reasoning poses a great challenge to LLMs.
Currently, there is an increasing interest in the reasoning ability of LLMs.
Evaluations include several perspectives, such as mathematical reasoning, commonsense reasoning, logical reasoning, and domain-specific reasoning~\cite{DBLP:journals/corr/abs-2307-03109,DBLP:journals/corr/abs-2304-06364,DBLP:journals/corr/abs-2302-04023,DBLP:journals/corr/abs-2310-09107}.
However, most current reasoning evaluations focus primarily on the accuracy of the final answer and neglect a comprehensive assessment of the reasoning process.
Such evaluation is not ideal for understanding the reasoning ability of models, as it ignores situations where models may obtain correct answers through unfaithful or spurious reasoning shortcuts~\cite{DBLP:conf/iclr/Saparov023}.
Some recent research has started to evaluate the validity of the intermediate reasoning steps of LLMs~\cite{golovneva2022roscoe,DBLP:journals/corr/abs-2304-10703}.
However, they mainly focus on the relationship between the intermediate step and the final answer rather than measuring whether the model understands the intermediate reasoning process.
This paper proposes a novel logical reasoning benchmark that requires intermediate counterfactual thinking under logical relation restrictions. This leads to a more in-depth study of the model's intermediate reasoning process.

\vspace{-10pt}
\section{\ourdataset\ Benchmark}
\label{sec:dataset}
\vspace{-5pt}
\subsection{Task Definition}
\vspace{-3pt}
The desideratum is to harvest LLM counterfactual thinking and then investigate the validation of the thinking and its alignment with human counterfactual thinking. 
To achieve this, the LLM should generate its counterfactual thinking under proper logical scenarios. 
We design a task of counterfactual modification of argument text given a perturbation of premise given a static logical relation.

A demonstration of the proposed \ourtask\ is shown in Figure~\ref{fig:task}. 
An LLM is given the instruction as shown on the left-hand side, which can be illustrated by the diagram on the right-hand side. 
In the given instruction, \textsf{Argument} and \pone\ are related by a logical relation. We consider four main relations in practice, which are (\textsf{R1}) the premise \textit{provides a necessary assumption to} the argument, (\textsf{R2}) the premise \textit{provides a sufficient assumption to} the argument, (\textsf{R3}) the premise \textit{strengthen} the argument, and (\textsf{R4}) the premise \textit{weaken} the argument. The \textsf{Argument} and \pone then constitute \textsf{State 1} of the logical relation \textsf{R}.
The additional \ptwo\ perturbs the logical relation \textsf{R}. 
The goal for the LLM is to maintain a \textsf{State 2} with the given \ptwo\ and a modified \textsf{Argument$^{\prime}$} that \textsf{R} stands. To this end, it should properly edit the statements in \textsf{Argument} until the goal is reached. 
\Blue{Table~\ref{tab:example} lists \textsf{R} types, proportions, and sample questions.}

\vspace{-5pt}
\subsection{Benchmark Construction}
\vspace{-5pt}
Given a data point\footnote{Data source in Appendix~\ref{app:source}.} with context (the argument text), question, options, and the correct answer option, an annotator is required to provide a chosen wrong option (as the \ptwo) and a corresponding modified context (i.e., the modified \textsf{Argument$^{\prime}$}) to form a data point. 
The annotator is first instructed to read the whole question and comprehend the in-line logical relations, then choose one of the wrong options. After that, he/she edits the context by deleting, adding, or replacing text spans in the statements. 
The number of editions and the length of edited text spans are unrestricted as long as the statement partition is maintained. 

We then post-process the question and the annotation so that for each data point, \textsf{Argument} comes from the original context, \pone\ comes from the correct option, and \ptwo\ and \argprime\ come from the annotation. 

\noindent\textbf{Annotation Verification Process}
The data construction process includes 3 phases.
\Blue{In the first phase, 10 annotators write the gold \argprime\ following the routine introduced above. 
In the second phase, the other 5 annotators manually check the written \argprime\ by scoring the logic
pairs $(\argprime, \ptwo)$ 1 if a pair meets the logical relation, otherwise 0. The pairs $(\argm,
\pone)$ and $(\argm, \ptwo)$ as two control groups. The $(\argprime, \ptwo)$ pairs scored 0
are returned to annotators in the first phase for revision.
In the third phase, we further invite an expert who has a Ph.D. degree in logic and rhetoric to
manually verify 30 randomly sampled annotations. 28 out of 30 are verified with certainty. Therefore,
we find that the data in \ourdataset\ is of high quality.} \vspace{-5pt}

% fig 2
\begin{figure}[t]
    \centering
    \vspace{-10pt}
    \includegraphics[width=.8\linewidth]{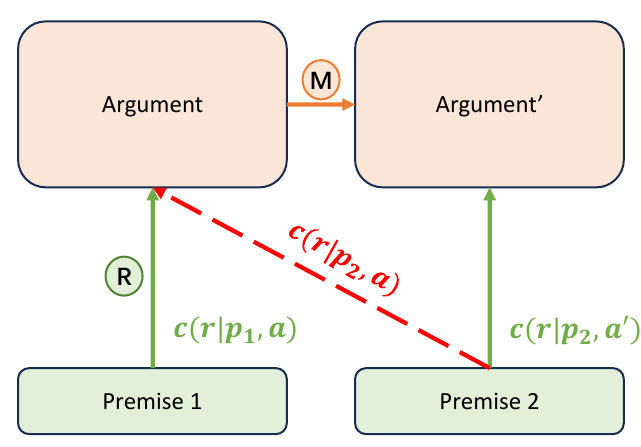}
    \vspace{-10pt}
    \caption{The concept graph of counterfactual logical modification.}
    \vspace{-10pt}
    \label{fig:concept_graph}
\end{figure}

\subsection{Data Statistics}
\vspace{-5pt}
\Blue{Tables~\ref{tab:stat_data} and \ref{tab:stat_token} demonstrate the dataset size, edit distance, and lengths of inputs of the gold inputs of \ourdataset.
\ourdataset\ contains 1,000 manually constructed high-quality data points.
According to Table~\ref{tab:stat_token}, the input prompts of the chain-of-thought setting (CoT) have a medium of 379 tokens in the training set. The zero-shot setting (Zero) has a medium of 368 tokens in prompts, while the medium token length in the few-shot setting (Few) is up to 1,328 in the training set.
Additionally, 
the output sequences Argument' are a modification of Argument. 
The edit distance statistics in Table~\ref{tab:stat_data} shows that the most challenging data point in \ourdataset\ has an edit distance of 66. The overall medium edit distance is 10, and
the test set medium edit distance is 13. Therefore, \ourdataset\ is a very challenging task.}

% table 2
\begin{table*}[t]
    \small
    \centering
    \vspace{-10pt}
    \begin{tabular}{
        l|rrrrrr|rrr
    }
        \toprule
                & \multicolumn{5}{c}{Dataset Size}  && \multicolumn{3}{c}{Edit Distance} \\
                % & \cline{1-5} && \cline{}
                & All & Necessary Assumption & Sufficient Assumption & Strength & Weaken && Min & Max & Medium \\
        \midrule
        Overall & 1,000 & 385 & 66 & 187 & 362      && 1 & 66 & 10 \\
        Train   & 600 & 227 & 37 & 118 & 218        && 1 & 65 & 8 \\
        Dev     & 200 & 79 & 15 & 34 & 72           && 1 & 60 & 13 \\
        Test    & 200 & 79 & 14 & 35 & 72           && 1 & 66 & 13 \\
        \bottomrule
    \end{tabular}
    \vspace{-5pt}
    \caption{
        \Blue{Statistics of \ourdataset.} 
    }
    \vspace{-10pt}
    \label{tab:stat_data}
\end{table*}

% table 3
\begin{table}[t]
    \small
    \centering
    \begin{tabular}{
        lrrrr
    }
        \toprule
                    & Max & Min & Mean & Medium \\
        \midrule
        CoT-Train   & 620 & 231 & 382.7 & 379 \\
        CoT-Dev     & 576 & 247 & 378.9 & 376 \\
        CoT-Test    & 620 & 216 & 374.7 & 371 \\
        \midrule
        Few-Train   & 1,569 & 1,180 & 1,331.7 & 1,328 \\
        Few-Dev     & 1,525 & 1,196 & 1,327.9 & 1,325 \\
        Few-Test    & 1,569 & 1,165 & 1,323.7 & 1,320 \\
        \midrule
        Zero-Train  & 609 & 220 & 371.7 & 368 \\
        Zero-Dev    & 565 & 236 & 367.9 & 365 \\
        Zero-Test   & 609 & 205 & 363.7 & 360 \\
        \bottomrule
    \end{tabular}
    \vspace{-5pt}
    \caption{
        \Blue{\ourdataset\ input length statistics by number of tokens.
        CoT: The chain-of-thought setting. 
        Few: The few-shot setting.
        Zero: The zero-shot setting.}
    }
    \vspace{-15pt}
    \label{tab:stat_token}
\end{table}

\vspace{-10pt}
% table 4: prompt. 3_cot
\begin{table*}[t]
    \small
    % \centering
    \vspace{-10pt}
    \begin{tabular}{
        |l|p{.85\textwidth}|
    }
        \hline
        \multirow{3}*{$c(r|p_1, a)$}
            & \texttt{You are an expert in logic. <definition of relation>. In the following, you are given an Argument and a Premise. Is the Premise <relation> the Argument? Please think step by step, and then answer ``yes'' or ``no''.} \\
            & \texttt{Argument: <$a$>} \\
            & \texttt{Premise: <$p_1$>} \\
        \hline
        \multirow{3}*{$c(r|p_2, a')$}
            & \texttt{You are an expert in logic. <definition of relation>. In the following, you are given an Argument and a Premise. Is the Premise <relation> the Argument? Please think step by step, and then answer ``yes'' or ``no''.} \\
            & \texttt{Argument: <$a'$>} \\
            & \texttt{Premise: <$p_2$>} \\
        \hline
        \multirow{3}*{$c(r|p_2, a)$}
            & \texttt{You are an expert in logic. <definition of relation>. In the following, you are given an Argument and a Premise. Is the Premise <relation> the Argument? Please think step by step, and then answer ``yes'' or ``no''.} \\
            & \texttt{Argument: <$a$>} \\
            & \texttt{Premise: <$p_2$>} \\
        \hline
    \end{tabular}
    \vspace{-5pt}
    \caption{
        Prompts for LLM decomposed evaluation task in SES.
    }
    \vspace{-5pt}
    \label{tab:prompt}
\end{table*}

% table 5
\begin{table*}[!t]
    \centering
    \small
    \begin{threeparttable}
        \begin{tabular}{
            lccccc
          }
            \toprule
                 & SES & SES$_\text{NA}$ & SES$_\text{SA}$ & SES$_\text{S}$ & SES$_\text{W}$ \\
            \midrule
            Human Performance                                   & 0.580 & 0.456 & 0.500 & 0.771 & \textbf{0.639} \\
            \hline \midrule
            GPT-3.5-Turbo \cite{gpt35}                          & 0.335 & 0.405 & 0.143 & 0.486 & 0.222 \\
            \midrule
            GPT-4 \cite{DBLP:journals/corr/abs-2303-08774}      & 0.475 & 0.544 & 0.643 & 0.714 & 0.250 \\
            \midrule
            GPT-4o\tnote{1}
                                                                & \textbf{0.680} & \textbf{0.696} & \textbf{0.667} & \textbf{0.800} & 0.625 \\    
            \bottomrule
        \end{tabular} 
        % \begin{tablenotes}
        %     \item [1] \url{https://openai.com/index/hello-gpt-4o/}
        % \end{tablenotes}
  \end{threeparttable}
  \vspace{-5pt}
  \caption{
    Self-evaluation scores (SES) of LLM-generated counterfactual statements. 
    \Blue{The backbone of SES is GPT-4.}
    NA: \textit{Necessary Assumption.}
    SA: \textit{Sufficient Assumption.}
    S: \textit{Strengthen.}
    W: \textit{Weaken.}
  }
  \vspace{-5pt}
  \label{tab:llm}
\end{table*}

% \vspace{-5pt}
\section{SES: Self-Evaluation Scores}
\vspace{-5pt}
We aim to use the proven complex reasoning capabilities of the large language model itself to perform faster and more efficient reasoning evaluations of complex reasoning tasks that do not have easy access to standard/human-tested answers. Specifically, we split the scenario of complex logical reasoning evaluation into several discriminative tasks that LLMs have already seen and have been heavily trained on through logical conceptual graphs for the model to perform high-precision reasoning evaluation. We then collect the evaluations of these simple tasks and compute the ratings of the complex reasoning tasks based on the structure of the logical concept graph.

\textbf{Counterfactual Modification Concept Graph} 
\Blue{Figure~\ref{fig:concept_graph} demonstrates the graph.}
To make the logic of counterfactual reasoning hold, we have a pair of primitive states $p_1$ (\pone) and $a$ (\argm) from $p_1$ and $a$ satisfying the relation $r$, i.e., $\Pr(r|p_1, a)$ approaches $1$.
The other claim $p_2$ (\ptwo) and the modified $a'$ (\argprime) satisfy the same relation $r$, i.e., $\Pr(r|p_2, a')$ approaches $1$.
In contrast, the relation between $p_2$ and the $a$ should not satisfy the relation $r$, i.e., $\Pr(r|p_2, a)$ approaches $0$.

\textbf{Decomposed Self-Evaluation Score}
As the demonstrated complex logical reasoning ability of current large language models, we let the large language model estimate $\Pr(r|p_1, a)$, $\Pr(r|p_2, a')$, and $\Pr(r|p_2, a)$, respectively. Specifically, we design a few binary classification tasks for the large language models, so that the probabilities are simplified to $c(r|p_1, a)$, $c(r|p_2, a')$, and $c(r|p_2, a)$ $\in\{0, 1\}$.
Table~\ref{tab:prompt} demonstrates the prompts.
%
% Finally, t
The overall logical modification score is computed by:
\vspace{-2pt}
\begin{equation}
    \small
    \vspace{-2pt}
    \begin{aligned}
        s = c(r|p_1, a)\times c(r|p_2, a') - c(r|p_2, a)\times c(r|p_2, a')
    \end{aligned}
    \vspace{-2pt}
    \label{eq}
\end{equation}

\noindent\Blue{The intuition of Eq.(\ref{eq}) is that, according to the concept graph in Figure~\ref{fig:concept_graph}, a desired Argument' results in:
i. 
The logical relation $r$ is satisfied in both pairs $(\ptwo, \argprime)$ and $(\pone, \argm)$, that is $\Pr(r|p_1, a) \times \Pr(r|p_2, a')$, where $\Pr(r|p_1, a)$ denotes the probability of logical relation $r$ holds given premise $p_1$ and argument $a$, and $\Pr(r|p_2, a')$ denote the probability of logical relation $r$ holds given the modified argument $a'$ and premise $p_2$. Practically, we prompt an LLM to classify the logical relation given an argument-premise pair and collect the responses $c(r|p_1, a)$ and $c(r|p_2, a')$. As a result, the first term in Eq.(\ref{eq}) is $c(r|p_1, a) \times c(r|p_2, a')$.
ii. 
The logical relation R in Figure~\ref{fig:concept_graph} can not hold between $(\ptwo, \argm)$. In other words, the probability of $r$ holds between $(\ptwo, \argm)$ should be distinguished from that between the modified $(\ptwo, \argprime)$ as much as possible, which is $-\Pr(r|p_2, a) \times \Pr(r|p_2, a')$. We use an LLM to do classification, resulting in $c(r|p_2, a) \times c(r|p_2, a')$ as the second term in Eq.(\ref{eq}).}

\vspace{-5pt}
\textbf{Alignment with Human Evaluation} 
We randomly select 50 samples from the test set and examined how well the self-evaluation score (SES) matches human evaluation. 
Specifically, we use GPT-4 to implement SES.
We recruit experts in argumentation to evaluate the modified arguments generated by GPT-4 on the selected 50 samples, scoring $1$ for good answers and $0$ for bad answers. The Cohen’s Kappa coefficient between the human annotators is $\kappa=0.6785$, indicating substantial consistency of human perspective. 
We then use the self-evaluation score to evaluate the same group of modified arguments again. 
Cohen's kappa coefficient between human and self-evaluation score is ${\kappa} = 0.4391$. 
This indicates that the self-evaluation score \Blue{is a safe reference and assistance for humans as the first study on automating the evaluation of the highly challenging counterfactual modification task.}
Therefore, we can \Blue{apply the SES score}
for more efficient logical modification evaluation.
\Blue{Also, we believe in a further improved automated score for this task, which we leave as a future work.}
\Blue{Additionally, the SES score can adjust to the latest and most cutting-edge models as backbones and it is convenient to check backbones' logical reasoning ability and alignment with humans.}

\vspace{-5pt}
\section{Experiments}
\label{sec:exp}
\vspace{-5pt}
\subsection{Main Results} % Section 5.1
\vspace{-5pt}
We first evaluate the large language models GPT-3.5-Turbo \cite{gpt35}, GPT-4 \cite{DBLP:journals/corr/abs-2303-08774}, and the latest GPT-4o\footnotemark[1].
% We first evaluate the large language models. We evaluate GPT-3.5-Turbo and GPT-4, respectively.
We also recruit 2 domain experts to contribute to the human performance. 
% \td{We also recruit experts in linguistics to annotate the answers with humans.}
Table~\ref{tab:llm} demonstrates the results. 
\Blue{The backbone of SES is GPT-4.}

% In the results, we see that t
The overall performance of the human experts on \ourdataset\ is 0.580, indicating that \ourdataset\ is quite a challenging task.
% In terms of the performance of t
Among the 4 logical relations, 
% \textit{strengthen} and \textit{weaken} are relatively easy, while 
\textit{necessary assumption} and \textit{sufficient assumption} are more challenging for humans.
\Blue{We consider a model to have counterfactual reasoning capabilities if its SES is comparable to, or even exceeds, those of humans.} 
% In addition, t
The performance of GPT-4 is slightly lower than that of humans, but it also demonstrates strong counterfactual logical reasoning ability.
And GPT-3.5-Turbo performs inferior to both human and GPT-4. 
We notice that GPT-4o exceeds human performance except that GPT-4o still has difficulty reasoning with the \textit{Weaken} relation.
Overall, large language models show great potential for counterfactual logical reasoning.

% fig 3
\begin{figure*}[!t]
    \centering
    \vspace{-10pt}
    \includegraphics[width=\textwidth]{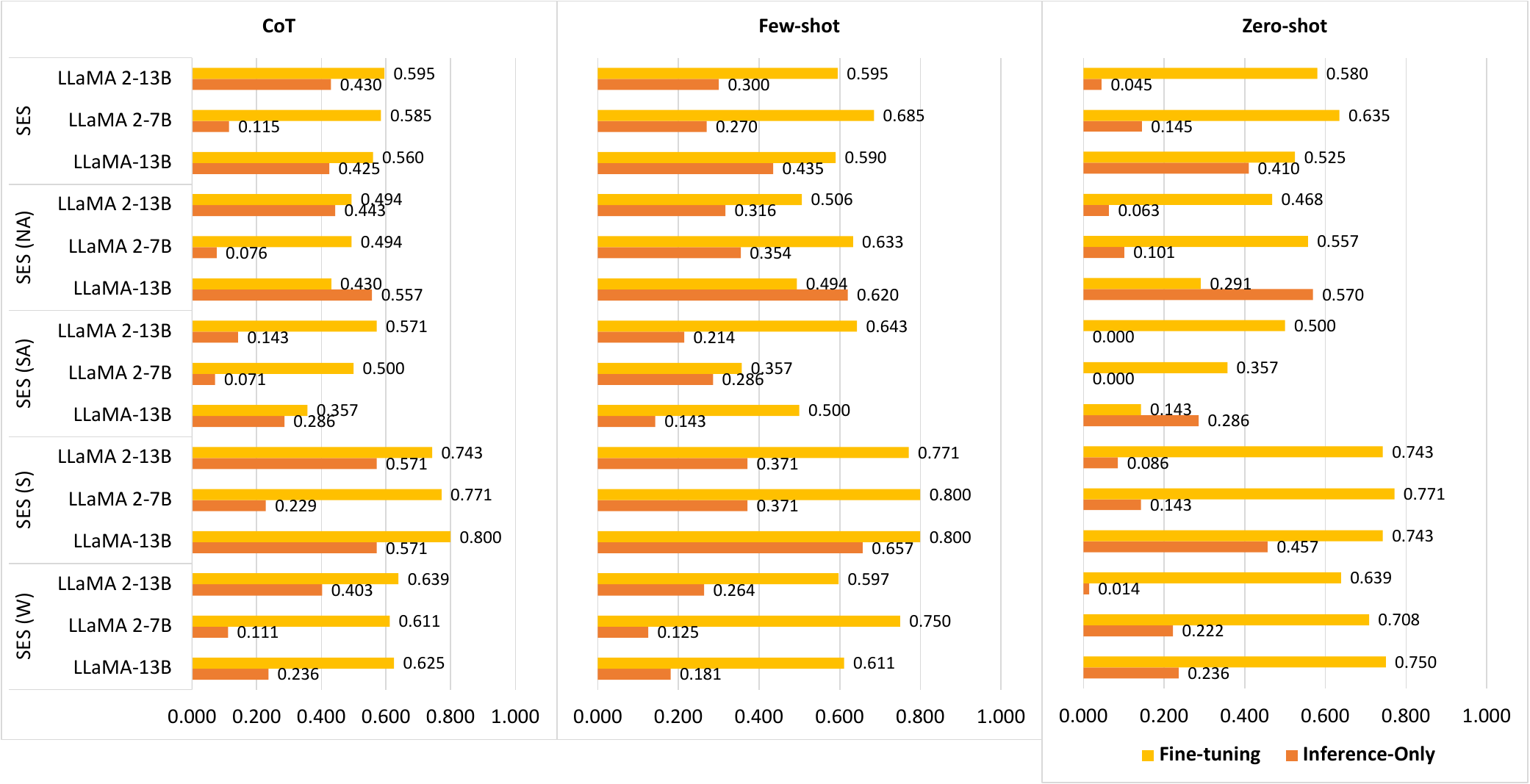}
    \vspace{-10pt}
    \caption{
        Per-relation performances of fine-tuned and inference-only LLaMA and LLaMA 2 families.
        NA: \textit{Necessary Assumption.}
        SA: \textit{Sufficient Assumption.}
        S: \textit{Strengthen.}
        W: \textit{Weaken.}    
        }
    \vspace{-5pt}
    \label{fig:per_rel}
\end{figure*}

% table 6
\begin{table*}[t]
    \small
    \centering
    \begin{tabular}{
        l|lll|lll|lll
    }
        \toprule
         && \multicolumn{2}{c}{CoT} && \multicolumn{2}{c}{Few} && \multicolumn{2}{c}{Zero} \\
        \textsf{R} 
         && w/o \textsf{R} & Full
         && w/o \textsf{R} & Full
         && w/o \textsf{R} & Full \\
        \midrule
        Necessary Assumption    && 0.434 & 0.494     && 0.052 & 0.633    && 0.234 & 0.557 \\
        Sufficient Assumption   && 0.286 & 0.500     && 0.000 & 0.357    && 0.234 & 0.557 \\
        Strengthen              && 0.457 & 0.771     && 0.114 & 0.800    && 0.343 & 0.771 \\
        Weaken                  && 0.222 & 0.611     && 0.014 & 0.750    && 0.243 & 0.708 \\
        \bottomrule
    \end{tabular}
    % \vspace{-5pt}
    \caption{
        \Blue{LLaMA 2-7B performances on test data with unseen logical relation. 
        w/o \textsf{R}: LLaMA 2-7B fine-tuned without \textsf{R}-type data.
        Full: LLaMA 2-7B fine-tuned with full training data.}
    }
    \vspace{-5pt}
    \label{tab:ablation}
\end{table*}

\vspace{-5pt}
\subsection{Fine-Tuning with Counterfactual Data}  % Section 5.3
\vspace{-5pt}
We then fine-tune LLaMA~\cite{DBLP:journals/corr/abs-2302-13971} and LLaMA 2~\cite{DBLP:journals/corr/abs-2307-09288}.
We randomly split the \ourdataset\ dataset into 60\%/20\%/20\% training/development/test data, and fine-tune the models with the \ourdataset\ training data. 
The implementation details are explained in Appendix~\ref{app:detail_finetune}.
We also compare the results with inference-only settings. 
The evaluation results are demonstrated in Figure~\ref{fig:per_rel}.

In general, the pre-trained LLaMA and LLaMA 2 models achieve a certain level of counterfactual reasoning, 
and fine-tuning with counterfactual data further improves the performance.
For example, LLaMA 2-13B with chain-of-thought prompting increases by 38.4\% (SES: 0.430 $\to$ 0.595),
with few-shot prompting increases by 98.3\% (SES: 0.300 $\to$ 0.595),
and with zero-shot prompting increases by 1,188.9\% (SES: 0.045 $\to$ 0.580).
The results indicate that such counterfactual data are barely seen in the LLaMA and LLaMA 2 pre-training data,
and datasets such as the proposed \ourdataset\ are much needed for developing models' counterfactual reasoning capabilities.

Moreover, among the four logical relations, 
\textit{Weaken} performances are significantly increased after fine-tuning.
However, 
the absolute SES scores of \textit{Sufficient Assumption} problems after fine-tuning are still relatively low.
It shows that \textit{Sufficient Assumption} are challenging.
Therefore, we still need a profound investigation of different logical relations to improve the models’ counterfactual logical reasoning ability.

\noindent\textbf{Ablation Study}
To study if there are only some \ourdataset\ training data is sufficient to reveal unseen logical relations, we further fine-tuned LLaMA-7b on training data excluding logical relation \textsf{R} (\textsf{R}=Necessary Assumption/Sufficient Assumption/Strengthen/Weaken), and evaluated it on test data that includes \textsf{R}. We then compared the performance of this model with itself trained on the full data set. Table~\ref{tab:ablation} shows that the performance of the unseen logical relation does drop drastically. This suggests that comprehensive learning across all types of logical relations is crucial. 
% \vspace{-2pt}

% table 7
\begin{table*}[!t]
\centering
\small
\vspace{-7pt}
  \begin{tabular}{
    lllccccc
  }
    \toprule    
    & Model & Params & SES & SES$_\text{NA}$ & SES$_\text{SA}$ & SES$_\text{S}$ & SES$_\text{W}$ \\    
    \midrule    
    \multirow{11}*{CoT}
    & Flan-T5-Large~\cite{DBLP:journals/corr/abs-2210-11416}      & 780M & 0.28 & 0.30 & \textbf{0.21} & 0.46 & 0.18 \\
    & Flan-T5-XL~\cite{DBLP:journals/corr/abs-2210-11416}         & 3B & 0.21 & 0.18 & 0.07 & 0.34 & 0.19 \\
    & Flan-T5-XXL~\cite{DBLP:journals/corr/abs-2210-11416}        & 11B & 0.26 & \textbf{0.33} & 0.07 & 0.37 & 0.17 \\
    & ChatGLM2-6B~\cite{du2022glm}                                & 6B & 0.25 & 0.32 & 0.07 & 0.34 & 0.17 \\
    & Baichuan2-7B-Chat~\cite{baichuan2023baichuan2}              & 7B & 0.24 & 0.20 & 0.07 & \textbf{0.49} & 0.18 \\
    & Baichuan2-13B-Chat~\cite{baichuan2023baichuan2}             & 13B & 0.26 & 0.19 & \textbf{0.21} & \textbf{0.49} & 0.22 \\    
    & InternLM-Chat-7B~\cite{2023internlm}                        & 7B & \textbf{0.31} & 0.32 & \textbf{0.21} & 0.43 & 0.25 \\
    & Vicuna-7B-v1.5~\cite{vicuna2023}                            & 7B & 0.15 & 0.13 & 0.00 & 0.29 & 0.14 \\
    & Vicuna-13B-v1.5~\cite{vicuna2023}                           & 13B & 0.26 & 0.18 & 0.14 & 0.46 & 0.28 \\
    & Qwen-14B-Chat~\cite{DBLP:journals/corr/abs-2309-16609}      & 14B & 0.30 & 0.24 & \textbf{0.21} & 0.46 & \textbf{0.29} \\    
    & WizardLM-13B-v1.2~\cite{xu2024wizardlm}                     & 13B & 0.02 & 0.04 & 0.00 & 0.00 & 0.00 \\
    \midrule
    \multirow{11}*{Few}
    & Flan-T5-Large~\cite{DBLP:journals/corr/abs-2210-11416}      & 780M & 0.16 & 0.11 & \textbf{0.07} & 0.29 & 0.17 \\
    & Flan-T5-XL~\cite{DBLP:journals/corr/abs-2210-11416}         & 3B & 0.16 & 0.09 & 0.00 & 0.29 & 0.19 \\
    & Flan-T5-XXL~\cite{DBLP:journals/corr/abs-2210-11416}        & 11B & 0.21 & 0.14 & \textbf{0.07} & \textbf{0.40} & \textbf{0.21} \\
    & ChatGLM2-6B~\cite{du2022glm}                                & 6B & 0.01 & 0.01 & 0.00 & 0.00 & 0.00 \\
    & Baichuan2-7B-Chat~\cite{baichuan2023baichuan2}              & 7B & 0.00 & 0.00 & 0.00 & 0.00 & 0.00 \\
    & Baichuan2-13B-Chat~\cite{baichuan2023baichuan2}             & 13B & 0.02 & 0.03 & 0.00 & 0.06 & 0.00 \\ 
    & InternLM-Chat-7B~\cite{2023internlm}                        & 7B & 0.01 & 0.00 & 0.00 & 0.00 & 0.01 \\
    & Vicuna-7B-v1.5~\cite{vicuna2023}                            & 7B & 0.01 & 0.01 & 0.00 & 0.00 & 0.00 \\
    & Vicuna-13B-v1.5~\cite{vicuna2023}                           & 13B & 0.01 & 0.01 & 0.00 & 0.00 & 0.00 \\
    & Qwen-14B-Chat~\cite{DBLP:journals/corr/abs-2309-16609}      & 14B & \textbf{0.25} & \textbf{0.29} & \textbf{0.07} & \textbf{0.40} & 0.17 \\ 
    & WizardLM-13B-v1.2~\cite{xu2024wizardlm}                     & 13B & 0.01 & 0.01 & 0.00 & 0.00 & 0.00 \\
    \midrule
    \multirow{11}*{Zero}
    & Flan-T5-Large~\cite{DBLP:journals/corr/abs-2210-11416}      & 780M & 0.24 & 0.29 & 0.07 & 0.37 & 0.15 \\
    & Flan-T5-XL~\cite{DBLP:journals/corr/abs-2210-11416}         & 3B & 0.25 & 0.22 & 0.07 & 0.43 & 0.22 \\
    & Flan-T5-XXL~\cite{DBLP:journals/corr/abs-2210-11416}        & 11B & 0.28 & 0.29 & \textbf{0.21} & 0.43 & 0.21 \\
    & ChatGLM2-6B~\cite{du2022glm}                                & 6B & 0.15 & 0.10 & 0.00 & 0.34 & 0.13 \\
    & Baichuan2-7B-Chat~\cite{baichuan2023baichuan2}              & 7B & 0.27 & 0.25 & 0.14 & 0.46 & 0.22 \\
    & Baichuan2-13B-Chat~\cite{baichuan2023baichuan2}             & 13B & 0.29 & 0.25 & \textbf{0.21} & 0.46 & 0.25 \\
    & InternLM-Chat-7B~\cite{2023internlm}                        & 7B & 0.28 & 0.28 & 0.14 & 0.43 & 0.22 \\
    & Vicuna-7B-v1.5~\cite{vicuna2023}                            & 7B & 0.22 & 0.16 & 0.14 & 0.46 & 0.17 \\
    & Vicuna-13B-v1.5~\cite{vicuna2023}                           & 13B & 0.25 & 0.18 & \textbf{0.21} & 0.40 & \textbf{0.26} \\
    & Qwen-14B-Chat~\cite{DBLP:journals/corr/abs-2309-16609}      & 14B & \textbf{0.30} & \textbf{0.32} & \textbf{0.21} & \textbf{0.49} & 0.21 \\
    & WizardLM-13B-v1.2~\cite{xu2024wizardlm}                     & 13B & 0.01 & 0.01 & 0.00 & 0.00 & 0.00 \\    
    \bottomrule
  \end{tabular}
  % \vspace{-5pt}
  \caption{
    Performances of smaller models in three inference-only settings.
    CoT: The chain-of-thought setting.
    Few: The few-shot setting.
    Zero: The zero-shot setting.
    More details are in Appendix~\ref{app:details}.
    NA: \textit{Necessary Assumption.}
    SA: \textit{Sufficient Assumption.}
    S: \textit{Strengthen.}
    W: \textit{Weaken.}    
  }
  \vspace{-7pt}
  \label{tab:smaller}
\end{table*}

% fig 4
\begin{figure*}[t]
\centering
    \vspace{-10pt}
    \includegraphics[width=\textwidth]{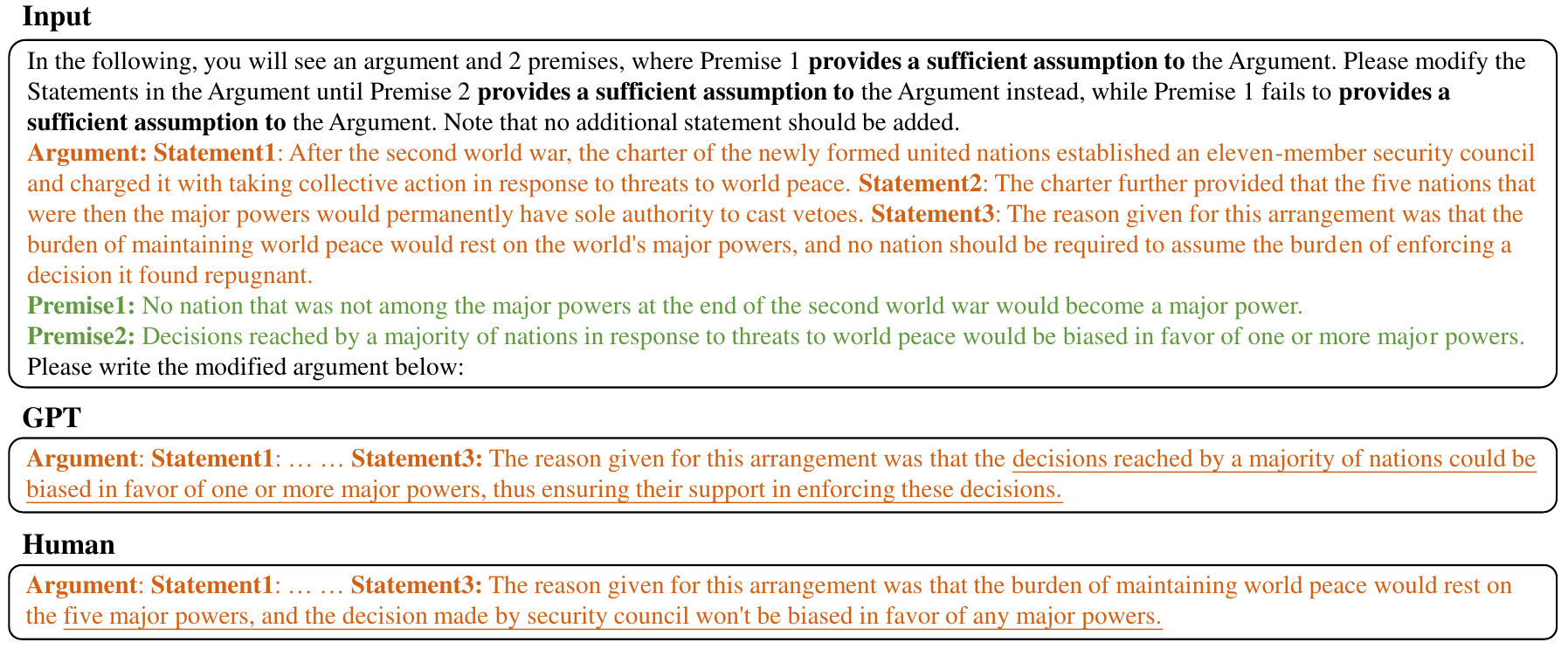}
    \vspace{-20pt}
    \caption{A successful case of counterfactual modification (reasoning type: \textit{Sufficient Assumption}) generated by GPT-4,
    which makes a logically consistent Argument$^{\prime}$ and is accordant with human reasoning.
    Counterfactually modified segments are \underline{underlined.}
    }
    \vspace{-7pt}
    \label{fig:case_good}
\end{figure*}

% fig 5
\vspace{-4pt}
\begin{figure*}[t]
\centering
    \vspace{-4pt}
    \includegraphics[width=\textwidth]{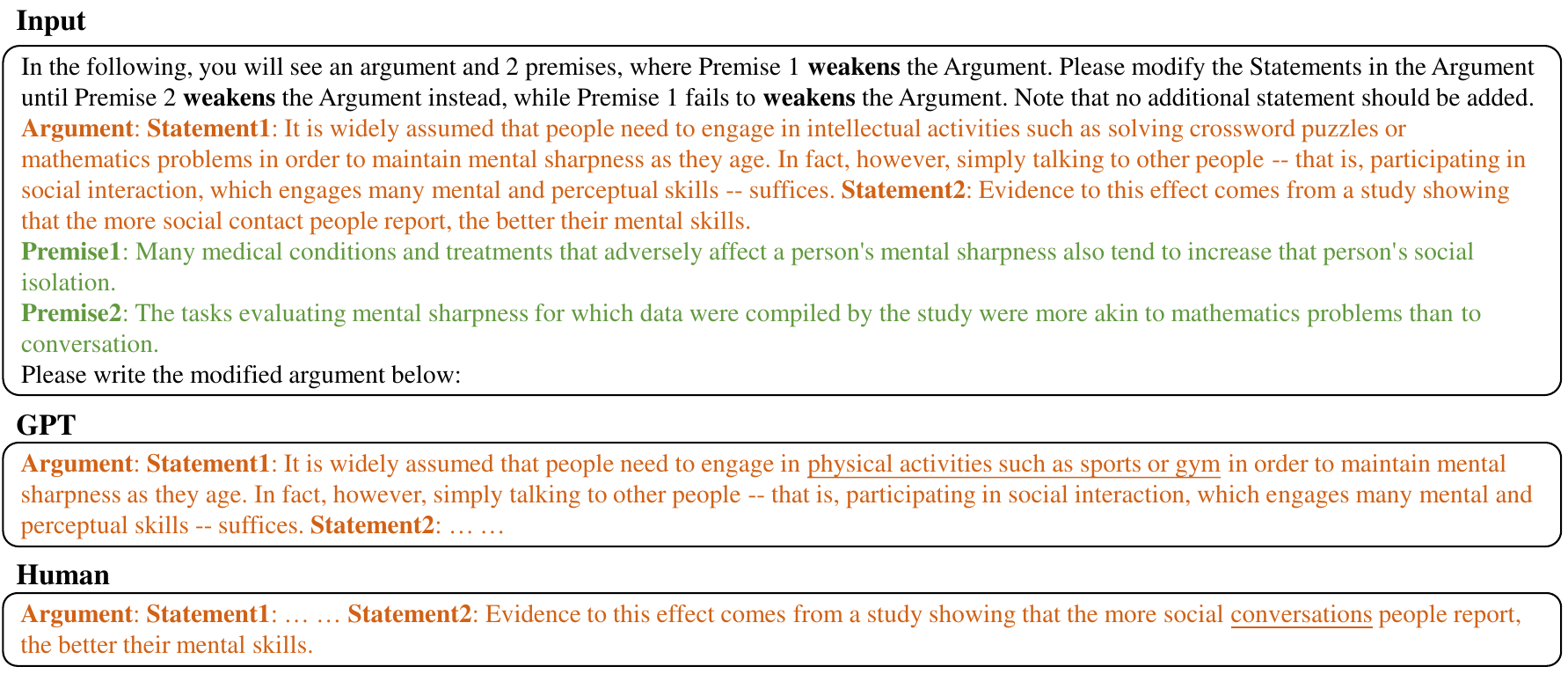}
    \vspace{-20pt}
    \caption{An inferior case of counterfactual modification (reasoning type: \textit{Weaken}) generated by GPT-4.
    It modifies Statement 1 by replacing intellectual activities with physical activities, where the logical restriction is not satisfied.
    Counterfactually modified segments are \underline{underlined.}
    }
    \vspace{-12pt}
    \label{fig:case_bad}
\end{figure*}

% \vspace{-3pt}
\subsection{Performances of Small Language Models}  % Section 5.2
\vspace{-4pt}
We further test various language models on smaller scales and the results are shown in Table~\ref{tab:smaller}.
All the models in Table~\ref{tab:smaller} directly perform inference without further fine-tuning thus examining their original counterfactual abilities. 
The detailed settings are demonstrated in Appendix~\ref{app:details},
and the brief introductions of the language models are listed in Appendix~\ref{app:slms}.
We have the following findings:
(1) Generally Speaking, all models perform inferiorly.
Some of the models, for example, Baichuan2-7B-Chat in the few-shot setting, hardly solve any of the counterfactual questions. 
Among the models, Qwen-14B-Chat performs the best in both the chain-of-thought setting and the zero-shot setting. 
But all models perform inferior to large language models or human
(2) The models perform better counterfactual reasoning with step-by-step reasoning processes (the CoT setting) while seeing more demonstrations in the prompt (the few-shot setting) harms the performances. 
It is indicated that the counterfactual cases have obscure reasoning patterns that are challenging for the models to transfer to unseen cases.
% It is indicated that the counterfactual reasoning task is not easily learned or transferred from similar cases. 
% It is not surprising since the counterfactual 
(3) The performances are correlated to the model scales, but the gaps are not necessarily significant. 
% For small language models, the performances are not necessarily correlated to model scale.
% in the angle of model scales
For instance, Flan-T5-XXL (11B) in general performs better than Flan-T5-Large (780M), and Vicuna-13B-v1.5 performs better than Vicuna-7B-v1.5. 
(4) For the four different relations, 
it is shown that \textit{Sufficient Assumption} and \textit{Weaken} are significantly more challenging than the other two reasoning types. 
The possible reason is that compared to the other two reasoning types, \textit{Sufficient Assumption} and \textit{Weaken} require more reasoning steps such as reversed thinking. 
Two cases are further shown in Section~\ref{sec:case}. 
To sum up, the \ourdataset\ task is challenging to current language models.
Therefore, further investigation is needed on the counterfactual reasoning abilities of language models.

\vspace{-4pt}
\subsection{Case Study}  % Section 5.4
\vspace{-5pt}
\label{sec:case}
We compare the modification by GPT-4 and humans. 
Figure~\ref{fig:case_good} shows a case about \textit{sufficient assumption}. That is, whether premise is a sufficient assumption for argument. 
The focus of the two premises here is to predict the impact of the charter provision.
Switching from Premise 1 to Premise 2, the focus of discussion changes from the group of major powers to other nations in response to threats to world peace. This mainly affects the elaboration of Statement 3. 
GPT-4 has revised Statement 3 accordingly. Humans also made changes to Statement 3. 
Both revisions emphasized the influence of the five major powers in Statement 3, which corresponded to Premise 2.
We find that GPT-4 can handle complex logical reasoning and counterfactual reasoning to a certain extent.

Figure~\ref{fig:case_bad} is about weakening an argument.
The argument is on human intellectual development. Statement 2 provides evidence to support Statement 1.
Premise 1 is on medical conditions and treatments providing counterexamples, and Premise 2 is on inaccuracies in research data. Human modifies Statement 2 to emphasize conversation, thus satisfying the logical conflict with the inaccuracies in research data described in Premise 2, thus satisfying the \textit{weaken} relation. 
GPT-4 modifies Statement 1 by replacing intellectual activities with physical activities such as sports or gym. 
Intellectual/physical activities have some counterfactual contrast, but in the context of the argument at hand, the logical relation restriction is not satisfied.
We find that GPT-4 can still be flawed in counterfactual reasoning.
In conclusion, complex counterfactual reasoning is challenging for large language models and needs improvements.

\vspace{-8pt}
\section{Conclusion}
\vspace{-8pt}
\label{sec:conclusion}
In this paper, we study large language models’ counterfactual reasoning capability under the constraint of proper logical relations. 
% Our goal is to harvest LLM counterfactual thinking and then investigate the validation of the thinking and its alignment with human counterfactual thinking. 
%
To this end, we introduce a novel task, \ourtask, that requires the LLMs to conduct counterfactual modification with logical restriction, where LLMs need to appropriately modify an argument text so that a specified logical relation stands. 
% This task not only challenges the LLMs but also lays the groundwork for more nuanced and logically coherent counterfactual reasoning.
%
To ensure a comprehensive evaluation, 
we then construct a benchmark dataset \ourdataset.
% by gathering human annotations and constructing a tailored dataset specifically designed for this purpose.
Moreover, we propose a 
% \textit{logic-aware counterfactual score} that quantifies and 
\textit{Self-Evaluation Score (SES)} that
decomposes the evaluation into several LLMs basic discrimination tasks, which is demonstrated aligned with human evaluations.
We further evaluate smaller language models in inference-only and fine-tuning manners.
The fine-tuned models' performances are significantly improved, but there is still a large gap with human performance.
% Furthermore, we evaluate small language models including the LLaMA and LLaMA 2 families on \ourdataset.
% The experimental results find that there is still room for LLMs to develop their counterfactual logical thinking.
Our findings thus underscore the need for further development in LLMs' counterfactual reasoning.

\vspace{-8pt}
\section{Limitations}
\vspace{-8pt}
The major limitation of this work is that we do not include multiple reference sentences in this version of \ourdataset. 
However, the proposed SES score leverages LLM to conduct a human-aligned evaluation, so it reduces the impact of reference in evaluating the model performances.

% \newpage
% \clearpage

\section*{Acknowledgements}
This work was supported in part by the National Science and Technology Major Project under Grant No. 2020AAA0109704,
% This work was supported in part by 
the Research Grants Council of the Hong Kong SAR under Grant GRF 11217823 and Collaborative Research Fund C1042-23GF, the National Natural Science Foundation of China under Grant 62371411, InnoHK initiative, the Government of the HKSAR, Laboratory for AI-Powered Financial Technologies.
The authors thank Dr. Xingchi Su, Guangyan Sun, and Cen Li for their great effort in carefully reviewing the data.

% Entries for the entire Anthology, followed by custom entries
\bibliography{anthology,custom}
\bibliographystyle{acl_natbib}

\appendix
% \vspace{-5pt}
\section{Ethical Considerations}
% \vspace{-7pt}
The data and annotations are collected without personal or confidential information. 
Therefore, there is no ethical concern to the best of our knowledge. 

\section{Samples from \ourdataset}
\label{app:samples}
Tables~\ref{tab:clomo_sample_1} and \ref{tab:clomo_sample_3} demonstrate examples from the \ourdataset\ dataset.

\section{Data Source}
\label{app:source}
Since applicable data for the proposed task is lacking, we build a benchmark dataset by carefully selecting argument texts and collecting human annotation of the modified \textsf{Argument$^{\prime}$}.
We choose to use ReClor~\cite{yu2020reclor} as the source data, considering that ReClor includes standardized multiple-choice questions on argument texts and logical relations from LSAT. We then recruit domain experts to conduct the annotation. 

\section{Implementation Details of Three Inference-Only Settings}  % Appendix A. 
\label{app:details}
Tables~\ref{tab:demonstration_few_type0} to \ref{tab:demonstration_cot_type3} demonstrate examples of three inference-only samples with input and output. 
% The three inference-only settings are the few-shot setting, zero-shot setting, and chain-of-thought settings. 
The three inference-only settings are:
\begin{itemize}
    \item \textbf{Few-shot setting:} We first give four demonstrations in the prompt, and then provide an unseen question for the language model. 
    The four demonstrations are randomly selected from the \ourdataset\ training set, each of which is from one of the four reasoning relations. 
    \item \textbf{Zero-shot setting:} We directly provide the question to be solved in the prompt.
    \item \textbf{Chain-of-thought setting:} We first provide the question, and then remind the language model to think step-by-step in the prompt.
\end{itemize}
Please see the Tables for the detailed demonstrations and prompts.

\section{Compared Models}
\label{app:slms}
% In this section, we give brief introductions of the language models used in the experiments. 
We use the following language models in the experiments. 

\textbf{Flan-T5}~\cite{DBLP:journals/corr/abs-2210-11416} is a family of language models that are instruction-finetuned on the T5 models~\cite{DBLP:journals/jmlr/RaffelSRLNMZLL20}.
The models perform well on commonsense reasoning, mathematics, history, law, and medicine. 
In this paper, we use Flan-T5-Large with 780M parameters, Flan-T5-XL with 3B parameters, and Flan-T5-XXL with 11B parameters.

\textbf{ChatGLM2}~\cite{du2022glm} is an open-source bilingual (Chinese-English) chat model based on GLM~\cite{DBLP:conf/acl/DuQLDQY022}. 
The models are refined with data with a context length of up to 32K. 
The model has strong performance on multiple reasoning tasks.
% It can also perform various reasoning. 

\textbf{Baichuan2}~\cite{baichuan2023baichuan2} is a family of language models trained on a high-quality corpus with 2.6 trillion tokens.
The models achieve good performance on multiple authoritative Chinese, English, and multi-language general and domain-specific benchmarks.
% The Chat models are further 
In the experiments, we use the Chat models Baichuan2-7B-Chat and Baichuan2-13B-Chat. 

\textbf{InternLM}~\cite{2023internlm} takes trillions of high-quality tokens for training to establish a powerful knowledge base.
It has outstanding comprehensive performance. 
We use the Chat model InternLM-Chat-7B in our experiments. 

\textbf{Vicuna-v1.5}~\cite{vicuna2023} models are fine-tuned on the LLaMA 2~\cite{DBLP:journals/corr/abs-2307-09288} models with supervised instruction fine-tuning with user-shared conversations collected from ShareGPT\footnote{\href{https://sharegpt.com/}{https://sharegpt.com/}}.

\textbf{Qwen}~\cite{DBLP:journals/corr/abs-2309-16609} is a series of comprehensive language model. 
The Qwen-Chat models are further fine-tuned with human alignment techniques such as Reinforcement Learning with Human Feedback (RLHF). 
The chat models have advanced tool use and planning capabilities.

\textbf{WizardLM}~\cite{xu2024wizardlm} is fine-tuned based on LLaMA~\cite{DBLP:journals/corr/abs-2302-13971} with a mixture of generated instruction data.
The model shows its benefits in various skills such as philosophy, technology, and physics.

\textbf{LLaMA}~\cite{DBLP:journals/corr/abs-2302-13971} is a collection of foundation language models trained on trillions of tokens with publicly available datasets. 

\textbf{LLaMA 2}~\cite{DBLP:journals/corr/abs-2307-09288} is a family of pre-trained and fine-tuned language models that can be adapted for a variety of natural language generation tasks.

\section{Implementation Details of Fine-Tuning}
% \section{Implementation Details of Fine-Tuning Models with CLoMo}
\label{app:detail_finetune}
We conduct full-parameter fine-tuning to LLaMA and LLaMA 2. 
The fine-tuning data is the \ourdataset\ training set.
Each model is fine-tuned by 10 epochs with a batch size of 4. 
The learning rate is $2e-5$ and is adapted by the cosine scheduler with a warmup proportion of $0.03$. 
% The models are evaluated on the \ourdataset\ development set every 10 steps. 
The best checkpoint is selected by the minimum perplexity in the validation split. 
In the inference phase, the models use beam search and a temperature of $0.7$.

% appdx table: dataset sample 0
% \vspace{-2pt}
\begin{table*}[!h] 
    \small
    \centering
    \vspace{-2pt}
    \begin{tabular}{
        >{\color{black}}p{\textwidth}
    }
        \toprule 
        \vspace{-4pt} \textbf{Input} \\
        \midrule \vspace{-3pt}
        In the following, you will see an argument and 2 premises, where Premise 1 
        \textbf{provides a necessary assumption} to the Argument. Please modify the Statements in the Argument until Premise 2 \textbf{provides a necessary assumption} to the Argument instead, while Premise 1 fails to \textbf{provides a necessary assumption} to the Argument. Note that no additional statement should be added. \\ \vspace{-5pt}

        \figOrange{
        \textbf{Argument:}
        \textbf{Statement1:} In 1963, a young macaque monkey was observed venturing into a hot spring to retrieve food which had fallen in. Soon, other macaques began to enter the spring, and over a few years this behavior was adopted by the entire troop. Prior to 1963, no macaques had ever been observed in the hot spring ; by 1990, the troop was regularly spending time there during the winters. 
        \textbf{Statement2:} Thus, these macaques are able to adopt and pass on new patterns of social behavior, and are not complete captives of their genetic heritage.
        } \\ \vspace{-5pt}
        
        \figGreen{
        \textbf{Premise1:} New patterns of behavior that emerge in macaque populations over the course of a few years or decades are not necessarily genetically predetermined.
        } \\ \vspace{-5pt}
        
        \figGreen{
        \textbf{Premise2:} The social behaviors of macaques are completely independent of their genetic heritage.
        } \\ \vspace{-5pt}
        
        {Please write the modified argument below:} \\ 
        
        \midrule        
        \vspace{-4pt} \textbf{Output}\\
        \midrule \vspace{-3pt}
        \figOrange{
        \textbf{Argument:}
        \textbf{Statement1:} In 1963, a young macaque monkey was observed venturing into a hot spring to retrieve food which had fallen in. Soon, other macaques began to enter the spring, and over a few years this behavior was adopted by the entire troop. Prior to 1963, no macaques had ever been observed in the hot spring ; by 1990, the troop was regularly spending time there during the winters.
        \textbf{Statement2:} Thus, these macaques\underline{' new patterns of social behavior must be determined by others.}
        } \\
        \bottomrule
    \end{tabular}
    \vspace{-7pt}
    \caption{
        \ourdataset\ data sample.
        Counterfactually modified segments are \underline{underlined.}
    }
    \vspace{-2pt}
    % "qtype": 0,
    % "id_string": "train_1754"
    \label{tab:clomo_sample_1}
\end{table*}

% appdx table: dataset sample 1
\vspace{-2pt}
\begin{table*}[!h] 
    \small
    \centering
    \vspace{-2pt}
    \begin{tabular}{
        >{\color{black}}p{\textwidth}
    }
        \toprule 
        \vspace{-4pt} \textbf{Input} \\
        \midrule \vspace{-3pt}
        In the following, you will see an argument and 2 premises, where Premise 1 
        \textbf{weakens} to the Argument. Please modify the Statements in the Argument until Premise 2 \textbf{weakens} to the Argument instead, while Premise 1 fails to \textbf{weakens} to the Argument. Note that no additional statement should be added. \\ \vspace{-5pt}

        \figOrange{
        \textbf{Argument:} 
        \textbf{Statement1:} Measurements of the motion of the planet uranus seem to show uranus being tugged by a force pulling it away from the sun and the inner planets. 
        \textbf{Statement2:} Neptune and pluto, the two known planets whose orbits are farther from the sun than is the orbit of uranus, do not have enough mass to exert the force that the measurements indicate. 
        \textbf{Statement3:} Therefore, in addition to the known planets, there must be at least one planet in our solar system that we have yet to discover.
        } \\ \vspace{-5pt}
        
        \figGreen{
        \textbf{Premise1:} There is a belt of comets beyond the orbit of pluto with powerful gravitational pull.
        } \\ \vspace{-5pt}
        
        \figGreen{
        \textbf{Premise2:} Neither neptune nor pluto is as massive as uranus.
        } \\ \vspace{-5pt}
        
        {Please write the modified argument below:} \\ 
        
        \midrule        
        \vspace{-4pt} \textbf{Output}\\
        \midrule \vspace{-3pt}
        \figOrange{
        \textbf{Argument:}
        \textbf{Statement1:} Measurements of the motion of the planet uranus seem to show uranus being tugged by a force pulling it away from the sun and the inner planets.
        \textbf{Statement2:} Neptune and pluto, the two known planets whose orbits are farther from the sun than is the orbit of uranus\underline{.  }
        \textbf{Statement3:} Therefore, \underline{one of the two planets must tug uranus.}
        } \\
        \bottomrule
    \end{tabular}
    \vspace{-7pt}
    \caption{
        \ourdataset\ data sample.
        Counterfactually modified segments are \underline{underlined.}
    }
    \vspace{-2pt}
    % "qtype": 3,
    % "id_string": "train_4341"
    \label{tab:clomo_sample_2}
\end{table*}

% appdx table: dataset sample 2
\vspace{-2pt}
\begin{table*}[!h] 
    \small
    \centering
    \vspace{-2pt}
    \begin{tabular}{
        >{\color{black}}p{\textwidth}
    }
        \toprule 
        \vspace{-4pt} \textbf{Input} \\
        \midrule \vspace{-3pt}
        In the following, you will see an argument and 2 premises, where Premise 1 
        \textbf{provides a necessary assumption} to the Argument. Please modify the Statements in the Argument until Premise 2 \textbf{provides a necessary assumption} to the Argument instead, while Premise 1 fails to \textbf{provides a necessary assumption} to the Argument. Note that no additional statement should be added. \\ \vspace{-5pt}

        \figOrange{
        \textbf{Argument:}        
        \textbf{Statement1:} Critic : historians purport to discover the patterns inherent in the course of events. But historians actually impose, rather than find, such patterns by choosing what to include in and exclude from their historical narratives. 
        \textbf{Statement2:} Thus, properly understood, histories reveal more about the presuppositions underlying different historians' attempts to understand what happened than about what actually happened.
        } \\ \vspace{-5pt}
        
        \figGreen{
        \textbf{Premise1:} Which pattern a historian imposes upon events is affected by that historian's presuppositions.
        } \\ \vspace{-5pt}
        
        \figGreen{
        \textbf{Premise2:} Historians have many presuppositions in common with one another.
        } \\ \vspace{-5pt}
        
        {Please write the modified argument below:} \\ 
        
        \midrule        
        \vspace{-4pt} \textbf{Output}\\
        \midrule \vspace{-3pt}
        \figOrange{
        \textbf{Argument:}
        \textbf{Statement1:} Critic : historians purport to discover the patterns inherent in the course of events. But historians actually impose, rather than find, such patterns by \underline{presupposing to choose} what to include in and exclude from their historical narratives. 
        \textbf{Statement2:} Thus, properly understood, histories reveal more about the presuppositions underlying different historians' attempts to understand what happened than about what actually happened\underline{, and these patterns are similar.}
        } \\
        \bottomrule
    \end{tabular}
    \vspace{-7pt}
    \caption{
        \ourdataset\ data sample. 
        Counterfactually modified segments are \underline{underlined.}
    }
    \vspace{-2pt}
    % "qtype": 0,
    % "id_string": "train_4580"
    \label{tab:clomo_sample_3}
\end{table*}

% appdx table 5. few.
\begin{table*}[!h] 
    \small
    \centering
    \begin{tabular}{
        >{\color{black}}p{\textwidth}
    }
        \toprule
        \textbf{Input:} \\
        In the following, you will see an argument and 2 premises, where Premise 1 \textbf{provides a necessary assumption to} the Argument. Please modify the Statements in the Argument until Premise 2 \textbf{provides a necessary assumption to} the Argument instead, while Premise 1 fails to \textbf{provides a necessary assumption to} the Argument. Note that no additional statement should be added. \\
        
        \textbf{Argument:}
        \textbf{Statement1:} Philosopher : both the consequences and the motives of human actions have bearing on the moral worth of those actions. 
        \textbf{Statement2:} Nonetheless, to be a moral agent one must have free will, because one can not be a moral agent without desiring to conform to a principle. 
        
        \textbf{Premise1:} Desiring to conform to a principle requires free will. 
        
        \textbf{Premise2:} It is impossible to have desires without also being a moral agent. \\
        
        Please write the modified argument below:
        
        \textbf{Modified Argument:}
        
        \textbf{Statement1:} Philosopher : both the consequences and the motives of human actions have bearing on the moral worth of those actions. \textbf{Statement2:} Nonetheless, one can not have desires without desiring to conform to a principle, because one can not be a moral agent without desiring to conform to a principle.\\
        
        \textbf{Argument:}
        \textbf{Statement1:} A reason larson can not do the assignment is that she has an unavoidable scheduling conflict. \textbf{Statement2:} On the other hand, a reason franks can not do the assignment is that he does not quite have the assertiveness the task requires. \textbf{Statement3:} So, the task must be assigned to parker, the only supervisor in the shipping department other than larson and franks. 
        
        \textbf{Premise1:} The task cannot be assigned to anyone other than a supervisor in the shipping department. 
        
        \textbf{Premise2:} The task cannot be assigned to anyone who has any kind of scheduling conflict.\\
        
        Please write the modified argument below:
        
        \textbf{Modified Argument:}
        
        \textbf{Statement1:} A reason larson can not do the assignment is that she has an unavoidable scheduling conflict. \textbf{Statement2:} On the other hand, a reason franks can not do the assignment is that he does not quite have the assertiveness the task requires. \textbf{Statement3:} So, the task must be assigned to parker, the only one who do not have scheduling conflict in the shipping department other than larson and franks.\\
        
        \textbf{Argument:}
        \textbf{Statement1:} Wart-gone cream is a well-known cure for reducing the spread of painful warts. A dime size amount rubbed on the infected area daily is the prescribed dosage, and applying the cream more often does not enhance the cream's healing effects. \textbf{Statement3:} Hence, distributing coupons giving \$ 3.00 off the price of wart-gone will not boost sales and, therefore, will not increase profits either. 
        
        \textbf{Premise1:} The \$3. 00 coupons, if distributed, would not induce new customers, who might not have purchased the product otherwise, to try wart-gone. 
        
        \textbf{Premise2:} Wart-gone can be produced in small or large batches without appreciably lowering the pre-tube production costs.\\
        
        Please write the modified argument below:
        
        \textbf{Modified Argument:}
        
        \textbf{Statement1:} Wart-gone cream is a well-known cure for reducing the spread of painful warts. A dime size amount rubbed on the infected area daily is the prescribed dosage, and applying the cream more often can enhance the cream's healing effects. \textbf{Statement3:} Hence, distributing coupons giving \$ 3.00 off the price of wart-gone will boost produce and, however, will not change pre-tube production costs much.\\
        
        \textbf{Argument:}
        \textbf{Statement1:} In defending the hyperion school of journalism from charges that its program is of little or no value to its students, the dean of the school pointed to its recent success in placing students : 65 percent of its graduates went on to internships or jobs in print or broadcast journalism. 
        
        \textbf{Premise1:} More than half of the school's students came from jobs in journalism to improve their skills. 
        
        \textbf{Premise2:} Some newspaper editors do not regard journalism schools as a necessary part of the training of a journalist.\\
        
        Please write the modified argument below:
        
        \textbf{Modified Argument:}
        
        \textbf{Statement1:} In defending the hyperion school of journalism from charges that its program is of little or no value to its students, the dean of the school pointed to its recent success in placing students : the journalism regards journalism schools as a necessary part of the training of a journalist.\\  % end demonstrations. table 5
        
        \textbf{Argument:}
        \textbf{Statement1:} Consumer advocate : there is no doubt that the government is responsible for the increased cost of gasoline, because the government's policies have significantly increased consumer demand for fuel, and as a result of increasing demand, the price of gasoline has risen steadily.
        
        \textbf{Premise1:} The government can bear responsibility for that which it indirectly causes.
        
        \textbf{Premise2:} Consumer demand for gasoline cannot increase without causing gasoline prices to increase.\\
        
        Please write the modified argument below: \\
        \midrule
        \textbf{Output:}\\
        \textbf{Statement1:} Consumer advocate : there is no doubt that the government partly leads to the increased cost of gasoline, because the government's policies have significantly increased consumer demand for fuel, and as a result of increasing demand, the price of gasoline has risen steadily undoubtedly. \\
        \bottomrule
    \end{tabular}
    \caption{
        % \td{few-shot. type=0.}
        Few-shot setting inference-only sample with input and output. 
        Logical relation: \textit{Necessary Assumption}.
    }
    % "qtype": 0,
    % "id_string": "train_3968"
    \label{tab:demonstration_few_type0}
\end{table*}

% appdx table 6. few.
\begin{table*}[!h] 
    \small
    \centering
    \vspace{-50pt}
    \begin{tabular}{
        >{\color{black}}p{\textwidth}
    }
        \toprule
        \textbf{Input:} \\
        In the following, you will see an argument and 2 premises, where Premise 1 \textbf{provides a sufficient assumption to} the Argument. Please modify the Statements in the Argument until Premise 2 \textbf{provides a sufficient assumption to} the Argument instead, while Premise 1 fails to \textbf{provides a sufficient assumption to} the Argument. Note that no additional statement should be added. \\

        \textbf{Argument:}
        \textbf{Statement1:} Philosopher : both the consequences and the motives of human actions have bearing on the moral worth of those actions. 
        \textbf{Statement2:} Nonetheless, to be a moral agent one must have free will, because one can not be a moral agent without desiring to conform to a principle. 
        
        \textbf{Premise1:} Desiring to conform to a principle requires free will. 
        
        \textbf{Premise2:} It is impossible to have desires without also being a moral agent. \\
        
        Please write the modified argument below:
        
        \textbf{Modified Argument:}
        
        \textbf{Statement1:} Philosopher : both the consequences and the motives of human actions have bearing on the moral worth of those actions. \textbf{Statement2:} Nonetheless, one can not have desires without desiring to conform to a principle, because one can not be a moral agent without desiring to conform to a principle.\\
        
        \textbf{Argument:}
        \textbf{Statement1:} A reason larson can not do the assignment is that she has an unavoidable scheduling conflict. \textbf{Statement2:} On the other hand, a reason franks can not do the assignment is that he does not quite have the assertiveness the task requires. \textbf{Statement3:} So, the task must be assigned to parker, the only supervisor in the shipping department other than larson and franks. 
        
        \textbf{Premise1:} The task cannot be assigned to anyone other than a supervisor in the shipping department. 
        
        \textbf{Premise2:} The task cannot be assigned to anyone who has any kind of scheduling conflict.\\
        
        Please write the modified argument below:
        
        \textbf{Modified Argument:}
        
        \textbf{Statement1:} A reason larson can not do the assignment is that she has an unavoidable scheduling conflict. \textbf{Statement2:} On the other hand, a reason franks can not do the assignment is that he does not quite have the assertiveness the task requires. \textbf{Statement3:} So, the task must be assigned to parker, the only one who do not have scheduling conflict in the shipping department other than larson and franks.\\
        
        \textbf{Argument:}
        \textbf{Statement1:} Wart-gone cream is a well-known cure for reducing the spread of painful warts. A dime size amount rubbed on the infected area daily is the prescribed dosage, and applying the cream more often does not enhance the cream's healing effects. \textbf{Statement3:} Hence, distributing coupons giving \$ 3.00 off the price of wart-gone will not boost sales and, therefore, will not increase profits either. 
        
        \textbf{Premise1:} The \$3. 00 coupons, if distributed, would not induce new customers, who might not have purchased the product otherwise, to try wart-gone. 
        
        \textbf{Premise2:} Wart-gone can be produced in small or large batches without appreciably lowering the pre-tube production costs.\\
        
        Please write the modified argument below:
        
        \textbf{Modified Argument:}
        
        \textbf{Statement1:} Wart-gone cream is a well-known cure for reducing the spread of painful warts. A dime size amount rubbed on the infected area daily is the prescribed dosage, and applying the cream more often can enhance the cream's healing effects. \textbf{Statement3:} Hence, distributing coupons giving \$ 3.00 off the price of wart-gone will boost produce and, however, will not change pre-tube production costs much.\\
        
        \textbf{Argument:}
        \textbf{Statement1:} In defending the hyperion school of journalism from charges that its program is of little or no value to its students, the dean of the school pointed to its recent success in placing students : 65 percent of its graduates went on to internships or jobs in print or broadcast journalism. 
        
        \textbf{Premise1:} More than half of the school's students came from jobs in journalism to improve their skills. 
        
        \textbf{Premise2:} Some newspaper editors do not regard journalism schools as a necessary part of the training of a journalist.\\
        
        Please write the modified argument below:
        
        \textbf{Modified Argument:}
        
        \textbf{Statement1:} In defending the hyperion school of journalism from charges that its program is of little or no value to its students, the dean of the school pointed to its recent success in placing students : the journalism regards journalism schools as a necessary part of the training of a journalist.\\  % end demonstrations. table 6
        
        \textbf{Argument:}
        \textbf{Statement1:} Super express shipping company has implemented a new distribution system that can get almost every package to its destination the day after it is sent. The company worries that this more efficient system will result in lower sales of its premium next-day delivery service, because its two day service will usually arrive the following day anyway. 
        \textbf{Statement2:} The company plans to encourage sales of its next-day service by intentionally delaying delivery of its two-day packages so that they will not be delivered the following day, even if the package arrives at its destination city in time for next-day delivery.
        
        \textbf{Premise1:} Deliberate delay of packages will not affect the company's image in a way that significantly reduces its ability to attract and retain customers.
        
        \textbf{Premise2:} The overnight service is too expensive to be attractive to most customers currently.\\
        
        Please write the modified argument below:\\
        \midrule
        \textbf{Output:}\\
        \textbf{Statement1:} Super express shipping company has paused to implemente a new distribution system that can get almost every package to its destination the day after it is sent. The company worries that this more efficient system will result in lower sales of its premium next-day delivery service, because its two day service will usually arrive the following day anyway. 
        \textbf{Statement2:} The company plans to encourage sales of its next-day service by stopping accelerating delivery of its two-day packages so that they will not be delivered the following day, leaving the package arriving at its destination city overnight for next-day delivery. \\
        \bottomrule
    \end{tabular}
    \caption{
        % \td{few-shot. type=1.}
        Few-shot setting inference-only sample with input and output. 
        Logical relation: \textit{Sufficient Assumption}.
    }
    % "qtype": 1,
    % "id_string": "train_4022"
    \label{tab:demonstration_few_type1}
\end{table*}

% appdx table 7. few.
\begin{table*}[!h] 
    \small
    \centering
    \vspace{-14mm}
    \begin{tabular}{
        >{\color{black}}p{\textwidth}
    }
        \toprule
        \textbf{Input:} \\
        In the following, you will see an argument and 2 premises, where Premise 1 \textbf{strengthens} the Argument. Please modify the Statements in the Argument until Premise 2 \textbf{strengthens} the Argument instead, while Premise 1 fails to \textbf{strengthens} the Argument. Note that no additional statement should be added. \\

        \textbf{Argument:}
        \textbf{Statement1:} Philosopher : both the consequences and the motives of human actions have bearing on the moral worth of those actions. 
        \textbf{Statement2:} Nonetheless, to be a moral agent one must have free will, because one can not be a moral agent without desiring to conform to a principle. 
        
        \textbf{Premise1:} Desiring to conform to a principle requires free will. 
        
        \textbf{Premise2:} It is impossible to have desires without also being a moral agent. \\
        
        Please write the modified argument below:
        
        \textbf{Modified Argument:}
        
        \textbf{Statement1:} Philosopher : both the consequences and the motives of human actions have bearing on the moral worth of those actions. \textbf{Statement2:} Nonetheless, one can not have desires without desiring to conform to a principle, because one can not be a moral agent without desiring to conform to a principle.\\
        
        \textbf{Argument:}
        \textbf{Statement1:} A reason larson can not do the assignment is that she has an unavoidable scheduling conflict. \textbf{Statement2:} On the other hand, a reason franks can not do the assignment is that he does not quite have the assertiveness the task requires. \textbf{Statement3:} So, the task must be assigned to parker, the only supervisor in the shipping department other than larson and franks. 
        
        \textbf{Premise1:} The task cannot be assigned to anyone other than a supervisor in the shipping department. 
        
        \textbf{Premise2:} The task cannot be assigned to anyone who has any kind of scheduling conflict.\\
        
        Please write the modified argument below:
        
        \textbf{Modified Argument:}
        
        \textbf{Statement1:} A reason larson can not do the assignment is that she has an unavoidable scheduling conflict. \textbf{Statement2:} On the other hand, a reason franks can not do the assignment is that he does not quite have the assertiveness the task requires. \textbf{Statement3:} So, the task must be assigned to parker, the only one who do not have scheduling conflict in the shipping department other than larson and franks.\\
        
        \textbf{Argument:}
        \textbf{Statement1:} Wart-gone cream is a well-known cure for reducing the spread of painful warts. A dime size amount rubbed on the infected area daily is the prescribed dosage, and applying the cream more often does not enhance the cream's healing effects. \textbf{Statement3:} Hence, distributing coupons giving \$ 3.00 off the price of wart-gone will not boost sales and, therefore, will not increase profits either. 
        
        \textbf{Premise1:} The \$3. 00 coupons, if distributed, would not induce new customers, who might not have purchased the product otherwise, to try wart-gone. 
        
        \textbf{Premise2:} Wart-gone can be produced in small or large batches without appreciably lowering the pre-tube production costs.\\
        
        Please write the modified argument below:
        
        \textbf{Modified Argument:}
        
        \textbf{Statement1:} Wart-gone cream is a well-known cure for reducing the spread of painful warts. A dime size amount rubbed on the infected area daily is the prescribed dosage, and applying the cream more often can enhance the cream's healing effects. \textbf{Statement3:} Hence, distributing coupons giving \$ 3.00 off the price of wart-gone will boost produce and, however, will not change pre-tube production costs much.\\
        
        \textbf{Argument:}
        \textbf{Statement1:} In defending the hyperion school of journalism from charges that its program is of little or no value to its students, the dean of the school pointed to its recent success in placing students : 65 percent of its graduates went on to internships or jobs in print or broadcast journalism. 
        
        \textbf{Premise1:} More than half of the school's students came from jobs in journalism to improve their skills. 
        
        \textbf{Premise2:} Some newspaper editors do not regard journalism schools as a necessary part of the training of a journalist.\\
        
        Please write the modified argument below:
        
        \textbf{Modified Argument:}
        
        \textbf{Statement1:} In defending the hyperion school of journalism from charges that its program is of little or no value to its students, the dean of the school pointed to its recent success in placing students : the journalism regards journalism schools as a necessary part of the training of a journalist.\\  % end demonstrations. table 7

        \textbf{Argument:}        
        \textbf{Statement1:} Often patients with ankle fractures that are stable, and thus do not require surgery, are given follow-up x-rays because their orthopedists are concerned about possibly having misjudged the stability of the fracture. 
        \textbf{Statement2:} When a number of follow-up x-rays were reviewed, however, all the fractures that had initially been judged stable were found to have healed correctly. 
        \textbf{Statement3:} Therefore, it is a waste of money to order follow-up x-rays of ankle fracture initially judged stable.
        
        \textbf{Premise1:} X-rays of patients of many different orthopedists working in several hospitals were reviewed.
        
        \textbf{Premise2:} Doctors who are general practitioners rather than orthopedists are less likely than orthopedists to judge the stability of an ankle fracture correctly.\\
        
        Please write the modified argument below: \\
        \midrule
        \textbf{Output:}\\
        \textbf{Statement1:} Often patients with ankle fractures that are stable, and thus do not require surgery, therefore, many general doctors do not order an x-ray to patients with stable ankle fractures in order to avoid wasting medical resources. 
        \textbf{Statement2:} When a number of follow-up x-rays were reviewed, however, most of the fractures that had initially been judged stable were found to haven't healed correctly. 
        \textbf{Statement3:} Therefore, it is necessary to order follow-up x-rays of ankle fracture initially judged stable. \\
        \bottomrule
    \end{tabular}
    \caption{
        % \td{few-shot. type=2.}
        Few-shot setting inference-only sample with input and output. 
        Logical relation: \textit{Strengthen}.
    }
    % "qtype": 2,
    % "id_string": "train_594"
    \label{tab:demonstration_few_type2}
\end{table*}

% appdx table 8. few.
\begin{table*}[!h] 
    \small
    \centering
    \vspace{-10mm}
    \begin{tabular}{
        >{\color{black}}p{\textwidth}
    }
        \toprule
        \textbf{Input:} \\
        In the following, you will see an argument and 2 premises, where Premise 1 \textbf{weakens} the Argument. Please modify the Statements in the Argument until Premise 2 \textbf{weakens} the Argument instead, while Premise 1 fails to \textbf{weakens} the Argument. Note that no additional statement should be added. \\
        
        \textbf{Argument:}
        \textbf{Statement1:} Philosopher : both the consequences and the motives of human actions have bearing on the moral worth of those actions. 
        \textbf{Statement2:} Nonetheless, to be a moral agent one must have free will, because one can not be a moral agent without desiring to conform to a principle. 
        
        \textbf{Premise1:} Desiring to conform to a principle requires free will. 
        
        \textbf{Premise2:} It is impossible to have desires without also being a moral agent. \\
        
        Please write the modified argument below:
        
        \textbf{Modified Argument:}
        
        \textbf{Statement1:} Philosopher : both the consequences and the motives of human actions have bearing on the moral worth of those actions. \textbf{Statement2:} Nonetheless, one can not have desires without desiring to conform to a principle, because one can not be a moral agent without desiring to conform to a principle.\\
        
        \textbf{Argument:}
        \textbf{Statement1:} A reason larson can not do the assignment is that she has an unavoidable scheduling conflict. \textbf{Statement2:} On the other hand, a reason franks can not do the assignment is that he does not quite have the assertiveness the task requires. \textbf{Statement3:} So, the task must be assigned to parker, the only supervisor in the shipping department other than larson and franks. 
        
        \textbf{Premise1:} The task cannot be assigned to anyone other than a supervisor in the shipping department. 
        
        \textbf{Premise2:} The task cannot be assigned to anyone who has any kind of scheduling conflict.\\
        
        Please write the modified argument below:
        
        \textbf{Modified Argument:}
        
        \textbf{Statement1:} A reason larson can not do the assignment is that she has an unavoidable scheduling conflict. \textbf{Statement2:} On the other hand, a reason franks can not do the assignment is that he does not quite have the assertiveness the task requires. \textbf{Statement3:} So, the task must be assigned to parker, the only one who do not have scheduling conflict in the shipping department other than larson and franks.\\
        
        \textbf{Argument:}
        \textbf{Statement1:} Wart-gone cream is a well-known cure for reducing the spread of painful warts. A dime size amount rubbed on the infected area daily is the prescribed dosage, and applying the cream more often does not enhance the cream's healing effects. \textbf{Statement3:} Hence, distributing coupons giving \$ 3.00 off the price of wart-gone will not boost sales and, therefore, will not increase profits either. 
        
        \textbf{Premise1:} The \$3. 00 coupons, if distributed, would not induce new customers, who might not have purchased the product otherwise, to try wart-gone. 
        
        \textbf{Premise2:} Wart-gone can be produced in small or large batches without appreciably lowering the pre-tube production costs.\\
        
        Please write the modified argument below:
        
        \textbf{Modified Argument:}
        
        \textbf{Statement1:} Wart-gone cream is a well-known cure for reducing the spread of painful warts. A dime size amount rubbed on the infected area daily is the prescribed dosage, and applying the cream more often can enhance the cream's healing effects. \textbf{Statement3:} Hence, distributing coupons giving \$ 3.00 off the price of wart-gone will boost produce and, however, will not change pre-tube production costs much.\\
        
        \textbf{Argument:}
        \textbf{Statement1:} In defending the hyperion school of journalism from charges that its program is of little or no value to its students, the dean of the school pointed to its recent success in placing students : 65 percent of its graduates went on to internships or jobs in print or broadcast journalism. 
        
        \textbf{Premise1:} More than half of the school's students came from jobs in journalism to improve their skills. 
        
        \textbf{Premise2:} Some newspaper editors do not regard journalism schools as a necessary part of the training of a journalist.\\
        
        Please write the modified argument below:
        
        \textbf{Modified Argument:}
        
        \textbf{Statement1:} In defending the hyperion school of journalism from charges that its program is of little or no value to its students, the dean of the school pointed to its recent success in placing students : the journalism regards journalism schools as a necessary part of the training of a journalist.\\  % end demonstrations. table 8

        \textbf{Argument:}
        \textbf{Statement1:} To suit the needs of corporate clients, advertising agencies have successfully modified a strategy originally developed for political campaigns. 
        \textbf{Statement2:} This strategy aims to provide clients with free publicity and air time by designing an advertising campaign that is controversial, thus drawing prime-time media coverage and evoking public comment by officials.
        
        \textbf{Premise1:} The usefulness of an advertising campaign is based solely on the degree to which the campaign's advertisements persuade their audiences.
        
        \textbf{Premise2:} Advertising agencies that specialize in campaigns for corporate clients are not usually chosen for political campaigns.\\
        
        Please write the modified argument below: \\
        \midrule
        \textbf{Output:}\\
        \textbf{Statement1:} To suit the needs of corporate clients, advertising agencies have successfully modified a strategy originally developed for political campaigns. 
        \textbf{Statement2:} This strategy aims to provide clients with free publicity and air time by designing an advertising campaign that is controversial, thus drawing prime-time media coverage and evoking public comment by officials. often such advertising agencies are chosen as partners for political campaigns.\\
        \bottomrule
    \end{tabular}
    \caption{
        % \td{few-shot. type=3.}
        Few-shot setting inference-only sample with input and output. 
        Logical relation: \textit{Weaken}.
    }
    % "qtype": 3,
    % "id_string": "train_316"
    \label{tab:demonstration_few_type3}
\end{table*}

% appdx table 9. zero.
\begin{table*}[!h] 
    \small
    \centering
    \begin{tabular}{
        >{\color{black}}p{\textwidth}
    }
        \toprule
        \textbf{Input:} \\
        In the following, you will see an argument and 2 premises, where Premise 1 \textbf{provides a necessary assumption to} the Argument. Please modify the Statements in the Argument until Premise 2 \textbf{provides a necessary assumption to} the Argument instead, while Premise 1 fails to \textbf{provides a necessary assumption to} the Argument. Note that no additional statement should be added. \\

        \textbf{Argument:}
        \textbf{Statement1:} Consumer advocate : there is no doubt that the government is responsible for the increased cost of gasoline, because the government's policies have significantly increased consumer demand for fuel, and as a result of increasing demand, the price of gasoline has risen steadily.
        
        \textbf{Premise1:} The government can bear responsibility for that which it indirectly causes.
        
        \textbf{Premise2:} Consumer demand for gasoline cannot increase without causing gasoline prices to increase. \\
        
        Please write the modified argument below: \\
        \midrule
        \textbf{Output:}\\
        \textbf{Statement1:} Consumer advocate : there is no doubt that the government partly leads to the increased cost of gasoline, because the government's policies have significantly increased consumer demand for fuel, and as a result of increasing demand, the price of gasoline has risen steadily undoubtedly.\\
        \bottomrule
    \end{tabular}
    \caption{
        % \td{zero-shot. type=0.}
        Zero-shot setting inference-only sample with input and output. 
        Logical relation: \textit{Necessary Assumption}.
    }
    % "qtype": 0,
    % "id_string": "train_3968"
    \label{tab:demonstration_zero_type0}
\end{table*}

% appdx table 10. zero.
\begin{table*}[!h] 
    \small
    \centering
    \begin{tabular}{
        >{\color{black}}p{\textwidth}
    }
        \toprule
        \textbf{Input:} \\
        In the following, you will see an argument and 2 premises, where Premise 1 \textbf{provides a sufficient assumption to} the Argument. Please modify the Statements in the Argument until Premise 2 \textbf{provides a sufficient assumption to} the Argument instead, while Premise 1 fails to \textbf{provides a sufficient assumption to} the Argument. Note that no additional statement should be added. \\

        \textbf{Argument:}
        \textbf{Statement1:} Super express shipping company has implemented a new distribution system that can get almost every package to its destination the day after it is sent. The company worries that this more efficient system will result in lower sales of its premium next-day delivery service, because its two day service will usually arrive the following day anyway. 
        \textbf{Statement2:} The company plans to encourage sales of its next-day service by intentionally delaying delivery of its two-day packages so that they will not be delivered the following day, even if the package arrives at its destination city in time for next-day delivery.
        
        \textbf{Premise1:} Deliberate delay of packages will not affect the company's image in a way that significantly reduces its ability to attract and retain customers.
        
        \textbf{Premise2:} The overnight service is too expensive to be attractive to most customers currently.\\
        
        Please write the modified argument below: \\
        \midrule
        \textbf{Output:}\\
        \textbf{Statement1:} Super express shipping company has paused to implemente a new distribution system that can get almost every package to its destination the day after it is sent. The company worries that this more efficient system will result in lower sales of its premium next-day delivery service, because its two day service will usually arrive the following day anyway. 
        \textbf{Statement3:} The company plans to encourage sales of its next-day service by stopping accelerating delivery of its two-day packages so that they will not be delivered the following day, leaving the package arriving at its destination city overnight for next-day delivery. \\
        \bottomrule
    \end{tabular}
    \caption{
        % \td{zero-shot. type=1.}
        Zero-shot setting inference-only sample with input and output. 
        Logical relation: \textit{Sufficient Assumption}.
    }
    % "qtype": 1,
    % "id_string": "train_4022"
    \label{tab:demonstration_zero_type1}
\end{table*}

% appdx table 11. zero.
\begin{table*}[!h] 
    \small
    \centering
    \begin{tabular}{
        >{\color{black}}p{\textwidth}
    }
        \toprule
        \textbf{Input:} \\
        In the following, you will see an argument and 2 premises, where Premise 1 \textbf{strengthens} the Argument. Please modify the Statements in the Argument until Premise 2 \textbf{strengthens} the Argument instead, while Premise 1 fails to \textbf{strengthens} the Argument. Note that no additional statement should be added. \\

        \textbf{Argument:} 
        \textbf{Statement1:} Often patients with ankle fractures that are stable, and thus do not require surgery, are given follow-up x-rays because their orthopedists are concerned about possibly having misjudged the stability of the fracture. 
        \textbf{Statement2:} When a number of follow-up x-rays were reviewed, however, all the fractures that had initially been judged stable were found to have healed correctly. 
        \textbf{Statement3:} Therefore, it is a waste of money to order follow-up x-rays of ankle fracture initially judged stable.
        
        \textbf{Premise1:} X-rays of patients of many different orthopedists working in several hospitals were reviewed.
        
        \textbf{Premise2:} Doctors who are general practitioners rather than orthopedists are less likely than orthopedists to judge the stability of an ankle fracture correctly.\\
        
        Please write the modified argument below: \\
        \midrule
        \textbf{Output:}\\
        \textbf{Statement1:} Often patients with ankle fractures that are stable, and thus do not require surgery, therefore, many general doctors do not order an x-ray to patients with stable ankle fractures in order to avoid wasting medical resources. 
        \textbf{Statement2:} When a number of follow-up x-rays were reviewed, however, most of the fractures that had initially been judged stable were found to haven't healed correctly. 
        \textbf{Statement3:} Therefore, it is necessary to order follow-up x-rays of ankle fracture initially judged stable.\\
        \bottomrule
    \end{tabular}
    \caption{
        % \td{zero-shot. type=2.}
        Zero-shot setting inference-only sample with input and output. 
        Logical relation: \textit{Strengthen}.
    }
    % "qtype": 2,
    % "id_string": "train_594"
    \label{tab:demonstration_zero_type2}
\end{table*}

% appdx table 12. zero.
\begin{table*}[!h] 
    \small
    \centering
    \begin{tabular}{
        >{\color{black}}p{\textwidth}
    }
        \toprule
        \textbf{Input:} \\
        In the following, you will see an argument and 2 premises, where Premise 1 \textbf{weakens} the Argument. Please modify the Statements in the Argument until Premise 2 \textbf{weakens} the Argument instead, while Premise 1 fails to \textbf{weakens} the Argument. Note that no additional statement should be added. \\

        \textbf{Argument:} 
        \textbf{Statement1:} To suit the needs of corporate clients, advertising agencies have successfully modified a strategy originally developed for political campaigns. \textbf{Statement2:} This strategy aims to provide clients with free publicity and air time by designing an advertising campaign that is controversial, thus drawing prime-time media coverage and evoking public comment by officials.
        
        \textbf{Premise1:} The usefulness of an advertising campaign is based solely on the degree to which the campaign's advertisements persuade their audiences.
        
        \textbf{Premise2:} Advertising agencies that specialize in campaigns for corporate clients are not usually chosen for political campaigns.\\
        
        Please write the modified argument below: \\
        \midrule
        \textbf{Output:}\\
        \textbf{Statement1:} To suit the needs of corporate clients, advertising agencies have successfully modified a strategy originally developed for political campaigns. 
        \textbf{Statement2:} This strategy aims to provide clients with free publicity and air time by designing an advertising campaign that is controversial, thus drawing prime-time media coverage and evoking public comment by officials. often such advertising agencies are chosen as partners for political campaigns.\\
        \bottomrule
    \end{tabular}
    \caption{
        % \td{zero-shot. type=3.}
        Zero-shot setting inference-only sample with input and output. 
        Logical relation: \textit{Weaken}.
    }
    % "qtype": 3,
    % "id_string": "train_316"
    \label{tab:demonstration_zero_type3}
\end{table*}

% appdx table 13. cot.
\begin{table*}[!h] 
    \small
    \centering
    \begin{tabular}{
        >{\color{black}}p{\textwidth}
    }
        \toprule
        \textbf{Input:} \\
        In the following, you will see an argument and 2 premises, where Premise 1 \textbf{provides a necessary assumption to} the Argument. Please modify the Statements in the Argument until Premise 2 \textbf{provides a necessary assumption to} the Argument instead, while Premise 1 fails to \textbf{provides a necessary assumption to} the Argument. Note that no additional statement should be added. \\

        \textbf{Argument:}
        \textbf{Statement1:} Consumer advocate : there is no doubt that the government is responsible for the increased cost of gasoline, because the government's policies have significantly increased consumer demand for fuel, and as a result of increasing demand, the price of gasoline has risen steadily.
        
        \textbf{Premise1:} The government can bear responsibility for that which it indirectly causes.
        
        \textbf{Premise2:} Consumer demand for gasoline cannot increase without causing gasoline prices to increase.\\
        
        Please write the modified argument below. Let's think step by step: \\
        \midrule
        \textbf{Output:}\\
        \textbf{Statement1:} Consumer advocate : there is no doubt that the government partly leads to the increased cost of gasoline, because the government's policies have significantly increased consumer demand for fuel, and as a result of increasing demand, the price of gasoline has risen steadily undoubtedly. \\
        \bottomrule
    \end{tabular}
    \caption{
        % \td{cot. type=0.}
        Chain-of-thought setting inference-only sample with input and output. 
        Logical relation: \textit{Necessary Assumption}.
    }
    % "qtype": 0,
    % "id_string": "train_3968"
    \label{tab:demonstration_cot_type0}
\end{table*}

% appdx table 14. cot.
\begin{table*}[!h] 
    \small
    \centering
    \begin{tabular}{
        >{\color{black}}p{\textwidth}
    }
        \toprule
        \textbf{Input:} \\
        In the following, you will see an argument and 2 premises, where Premise 1 \textbf{provides a sufficient assumption to} the Argument. Please modify the Statements in the Argument until Premise 2 \textbf{provides a sufficient assumption to} the Argument instead, while Premise 1 fails to \textbf{provides a sufficient assumption to} the Argument. Note that no additional statement should be added.\\

        \textbf{Argument:}
        \textbf{Statement1:} Super express shipping company has implemented a new distribution system that can get almost every package to its destination the day after it is sent. The company worries that this more efficient system will result in lower sales of its premium next-day delivery service, because its two day service will usually arrive the following day anyway. 
        \textbf{Statement2:} The company plans to encourage sales of its next-day service by intentionally delaying delivery of its two-day packages so that they will not be delivered the following day, even if the package arrives at its destination city in time for next-day delivery.
        
        \textbf{Premise1:} Deliberate delay of packages will not affect the company's image in a way that significantly reduces its ability to attract and retain customers.
        
        \textbf{Premise2:} The overnight service is too expensive to be attractive to most customers currently. \\
        
        Please write the modified argument below. Let's think step by step: \\
        \midrule
        \textbf{Output:}\\
        \textbf{Statement1:} Super express shipping company has paused to implemente a new distribution system that can get almost every package to its destination the day after it is sent. The company worries that this more efficient system will result in lower sales of its premium next-day delivery service, because its two day service will usually arrive the following day anyway. 
        \textbf{Statement3:} The company plans to encourage sales of its next-day service by stopping accelerating delivery of its two-day packages so that they will not be delivered the following day, leaving the package arriving at its destination city overnight for next-day delivery. \\
        \bottomrule
    \end{tabular}
    \caption{
        % \td{cot. type=1.}
        Chain-of-thought setting inference-only sample with input and output. 
        Logical relation: \textit{Sufficient Assumption}.
    }
    % "qtype": 1,
    % "id_string": "train_4022"
    \label{tab:demonstration_cot_type1}
\end{table*}

% appdx table 15. cot.
\begin{table*}[!h] 
    \small
    \centering
    \begin{tabular}{
        >{\color{black}}p{\textwidth}
    }
        \toprule
        \textbf{Input:} \\
        In the following, you will see an argument and 2 premises, where Premise 1 stren\textbf{}gthens the Argument. Please modify the Statements in the Argument until Premise 2 \textbf{strengthens} the Argument instead, while Premise 1 fails to \textbf{strengthens} the Argument. Note that no additional statement should be added. \\

        \textbf{Argument:} 
        \textbf{Statement1:} Often patients with ankle fractures that are stable, and thus do not require surgery, are given follow-up x-rays because their orthopedists are concerned about possibly having misjudged the stability of the fracture. 
        \textbf{Statement2:} When a number of follow-up x-rays were reviewed, however, all the fractures that had initially been judged stable were found to have healed correctly. 
        \textbf{Statement3:} Therefore, it is a waste of money to order follow-up x-rays of ankle fracture initially judged stable.
        
        \textbf{Premise1:} X-rays of patients of many different orthopedists working in several hospitals were reviewed.
        
        \textbf{Premise2:} Doctors who are general practitioners rather than orthopedists are less likely than orthopedists to judge the stability of an ankle fracture correctly.\\
        
        Please write the modified argument below. Let's think step by step: \\
        \midrule
        \textbf{Output:}\\
        \textbf{Statement1:} Often patients with ankle fractures that are stable, and thus do not require surgery, therefore, many general doctors do not order an x-ray to patients with stable ankle fractures in order to avoid wasting medical resources. 
        \textbf{Statement2:} When a number of follow-up x-rays were reviewed, however, most of the fractures that had initially been judged stable were found to haven't healed correctly. 
        \textbf{Statement3:} Therefore, it is necessary to order follow-up x-rays of ankle fracture initially judged stable. \\
        \bottomrule
    \end{tabular}
    \caption{
        % \td{cot. type=2.}
        Chain-of-thought setting inference-only sample with input and output. 
        Logical relation: \textit{Strengthen}.
    }
    % "qtype": 2,
    % "id_string": "train_594"
    \label{tab:demonstration_cot_type2}
\end{table*}

% appdx table 16. cot.
\begin{table*}[!h] 
    \small
    \centering
    \begin{tabular}{
        >{\color{black}}p{\textwidth}
    }
        \toprule
        \textbf{Input:} \\
        In the following, you will see an argument and 2 premises, where Premise 1 \textbf{weakens} the Argument. Please modify the Statements in the Argument until Premise 2 \textbf{weakens} the Argument instead, while Premise 1 fails to \textbf{weakens} the Argument. Note that no additional statement should be added. \\

        \textbf{Argument:} 
        \textbf{Statement1:} To suit the needs of corporate clients, advertising agencies have successfully modified a strategy originally developed for political campaigns. \textbf{Statement2:} This strategy aims to provide clients with free publicity and air time by designing an advertising campaign that is controversial, thus drawing prime-time media coverage and evoking public comment by officials.
        
        \textbf{Premise1:} The usefulness of an advertising campaign is based solely on the degree to which the campaign's advertisements persuade their audiences.
        
        \textbf{Premise2:} Advertising agencies that specialize in campaigns for corporate clients are not usually chosen for political campaigns.\\
        
        Please write the modified argument below. Let's think step by step: \\
        \midrule
        \textbf{Output:}\\
        \textbf{Statement1:} To suit the needs of corporate clients, advertising agencies have successfully modified a strategy originally developed for political campaigns. 
        \textbf{Statement2:} This strategy aims to provide clients with free publicity and air time by designing an advertising campaign that is controversial, thus drawing prime-time media coverage and evoking public comment by officials .often such advertising agencies are chosen as partners for political campaigns.\\
        \bottomrule
    \end{tabular}
    \caption{
        % \td{cot. type=3.}
        Chain-of-thought setting inference-only sample with input and output. 
        Logical relation: \textit{Weaken}.
    }
    % "qtype": 3,
    % "id_string": "train_316"
    \label{tab:demonstration_cot_type3}
\end{table*}

\end{document}